\documentclass[runningheads]{llncs}
\usepackage[T1]{fontenc}
\usepackage{graphicx}
\usepackage{lmodern}
\usepackage{multirow}
\usepackage{caption}
\usepackage{subcaption}
\usepackage{hyperref}
\usepackage{algorithm}
\usepackage{algorithmic}
\usepackage[nolist,nohyperlinks]{acronym}
\usepackage{soul}
\usepackage{xcolor}
\usepackage{amsmath}
\usepackage{amssymb}
\usepackage{booktabs}
\usepackage{tabularx}
\usepackage{array}
\usepackage{placeins}
\usepackage{makecell}    
\hypersetup{
    colorlinks=true,
    linkcolor=blue,
    citecolor=blue,
    urlcolor=blue
}
\usepackage{todonotes}

\begin{document}

\title{From Gradient Clipping to Structural Refinement: Improving DPSGD For Medical Image Segmentation}
\titlerunning{Improving DPSGD For Medical Image Segmentation}

\author{Shiva Parsarad, Parth Shandilya \and
Isabel Wagner
}
%
%
\institute{University of Basel\\
\email{shiva.parsarad@unibas.ch}\\
\email{parth.shandilya@stud.unibas.ch}\\
\email{isabel.wagner@unibas.ch}}
\maketitle              
\begin{abstract}
Medical image segmentation is widely used for disease detection but relies on sensitive data, raising privacy concerns as trained models can leak information. 
Differential privacy, typically implemented via \ac{dpsgd}, provides a solution, though at the cost of reduced utility.
Recent \ac{dpsgd} variants, including \ac{Auto-S}, \ac{NSGD}, and \ac{PSAC}, have shown promise in image classification, but their behavior in medical segmentation remains underexplored. 
We evaluate these methods across binary and multi-class tasks and analyze gradient alignment, showing that prior assumptions, particularly for \ac{PSAC}, do not consistently hold. 
We further demonstrate that combining clipping strategies with morphological refinement improves segmentation quality under privacy constraints. 
Finally, we propose an adaptive DP-Morph variant that captures class-specific structures and enhances performance in multi-class settings.

\end{abstract}
\keywords{Differential Privacy, Medical Segmentation, DP-Morph, Clipping strategies}

\begin{acronym}
  \acro{dpsgd}[DPSGD]{Differential Private Stochastic Gradient Descent}  
  \acro{dp}[DP]{Differentially Privacy}
  \acro{oct}[OCT]{Optical Coherence Tomography}
  \acro{PSAC}[PSAC]{Per-sample adaptive clipping}
  \acro{Auto-S}[Auto-S]{Automatic clipping}
  \acro{NSGD}[NSGD]{Normalised SGD with perturbation}
  \acro{HD95}[HD95]{95th percentile Hausdorff Distance}
  \acro{sgd}[SGD]{Stochastic Gradient Descent}
  \acro{ct}[CT]{Computed Tomography}
  \acro{SDMs}[SDMs]{signed distance maps}
  \acro{se}[$S_{\alpha}$]{Structure Measure}
  \acro{ee}[$E_{\phi}$]{Enhanced-Alignment Measure}

\end{acronym}

\section{Introduction}

Medical image segmentation relies on sensitive patient data, raising significant privacy concerns that cannot be fully addressed by traditional anonymization techniques \cite{liu2023surveya}.
In this context, \ac{dp} 
has emerged as the mathematical gold standard for privacy preservation, providing a provable, quantifiable guarantee that the presence or absence of any single individual in a dataset does not significantly alter the output of the resulting model.
Within the domain of deep learning, this is most commonly operationalized through \acf{dpsgd}, a mechanism that modifies \ac{sgd} to bound the influence of individual samples on the trained model \cite{abadi2016deep}.

Despite its strong guarantees, \ac{dpsgd} often suffers from substantial utility degradation due to gradient clipping and noise addition. 
The choice of clipping strategy plays a critical role in this trade-off, motivating recent alternatives to standard flat clipping, including normalization-based and adaptive methods such as \ac{Auto-S}, \ac{NSGD}, and \ac{PSAC} \cite{bu2023automaticc,yang2022normalizeda,xia2023differentiallya}.
However, these methods (e.g., \ac{Auto-S}, \ac{NSGD}, and \ac{PSAC}) have primarily been studied in classification settings and have not been evaluated in medical image segmentation, where dense and structured predictions introduce additional challenges.
Prior work suggests that normalization-based methods, such as \ac{Auto-S} and \ac{NSGD}, may introduce deviations between the effective update and the true gradient, particularly when per-sample gradients are small, while \ac{PSAC} aims to mitigate this effect \cite{xia2023differentiallya}. 
However, these claims have not been validated in segmentation tasks.

Beyond optimization, structural refinement has been proposed to mitigate performance degradation under \ac{dpsgd}. 
In particular, DP-Morph introduces morphological operations to recover structural details in segmentation outputs \cite{parsarad2025dpmorph}.
However, its interaction with alternative clipping strategies remains unexplored, and its use of uniform operations may be suboptimal for multi-class segmentation, where structures show distinct characteristics and may require class-specific refinement.

Motivated by these challenges, we study \ac{dpsgd} for medical image segmentation from two perspectives: clipping strategies and structural refinement.
On the optimization side, we evaluate alternative clipping methods (\ac{Auto-S}, \ac{NSGD}, and \ac{PSAC}) and analyze their gradient behavior.
On the structural side, we investigate their interaction with DP-Morph and propose a novel adaptive, class-specific extension for multi-class segmentation.
We investigate the following research questions:
(i) How do different \ac{dpsgd} clipping strategies behave in medical image segmentation across binary and multi-class tasks?
(ii) Do the gradient-distortion claims made for normalization-based clipping methods in prior work also hold in image segmentation, and does \ac{PSAC} mitigate this effect in practice?
(iii) How does morphological refinement (DP-Morph) interact with clipping strategies under privacy constraints?
(iv) Can adaptive, class-specific morphology improve performance in multi-class segmentation?

To answer these questions, we conduct a systematic evaluation on OCT (Duke, UMN) and CT (COVID-19) datasets, where Duke is a multi-class task and UMN and COVID-19 are binary tasks.
Our results show that conclusions from classification do not consistently transfer to segmentation, particularly for normalization-based methods.
We further demonstrate that morphological refinement improves performance under privacy constraints.
Finally, we propose an adaptive DP-Morph framework that achieves up to 1-2\% improvement in Dice.


\begingroup
\setlength{\textfloatsep}{8pt}
\begin{figure}[!t]
\centering

\begin{subfigure}[t]{0.32\linewidth}
    \centering
    \includegraphics[width=0.48\linewidth,height=2cm]{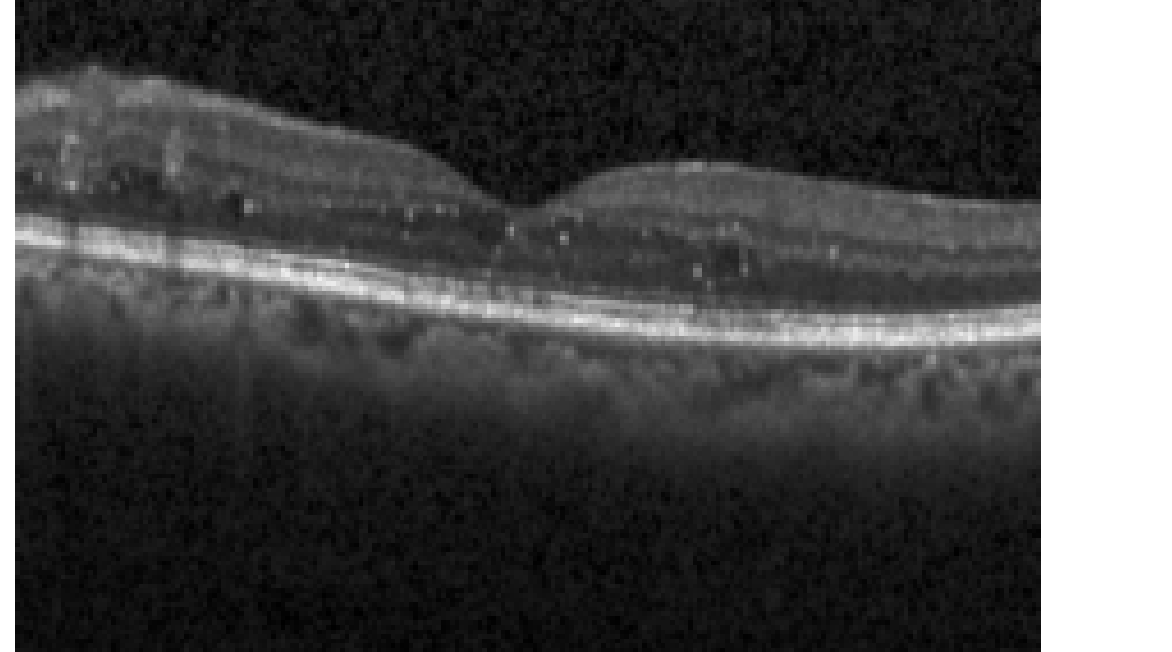}
    \hfill
    \includegraphics[width=0.48\linewidth,height=2cm]{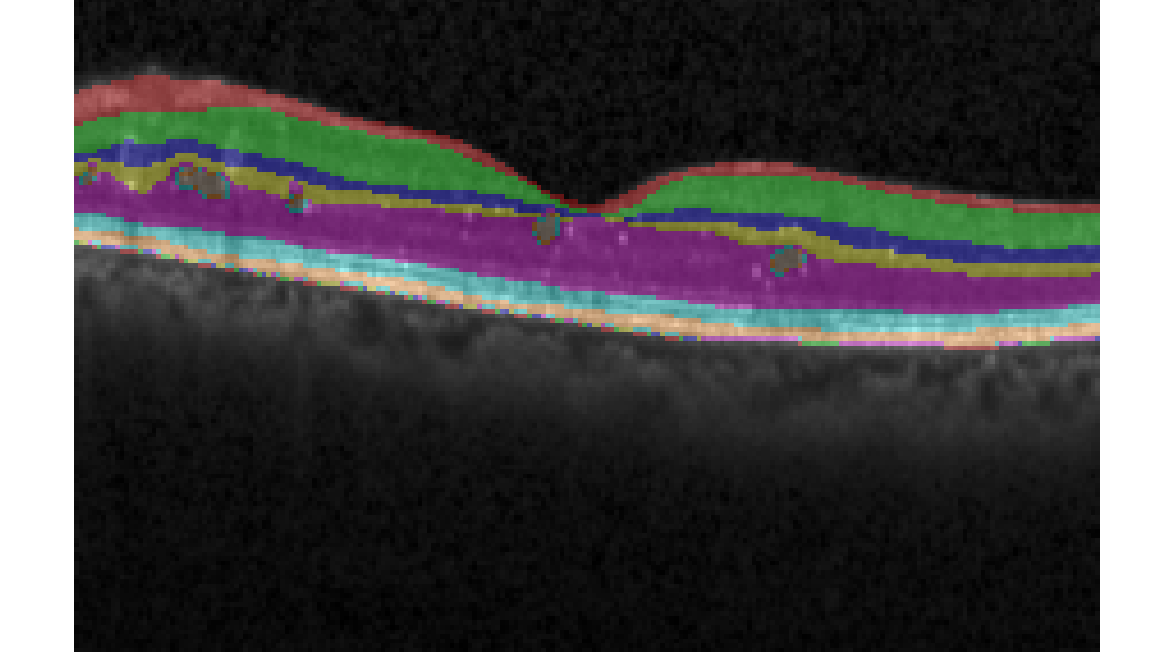}
    \caption{Duke}
    \label{fig:medical_duke}
\end{subfigure}
\hfill
\begin{subfigure}[t]{0.32\linewidth}
    \centering
    \includegraphics[width=0.48\linewidth,height=2cm]{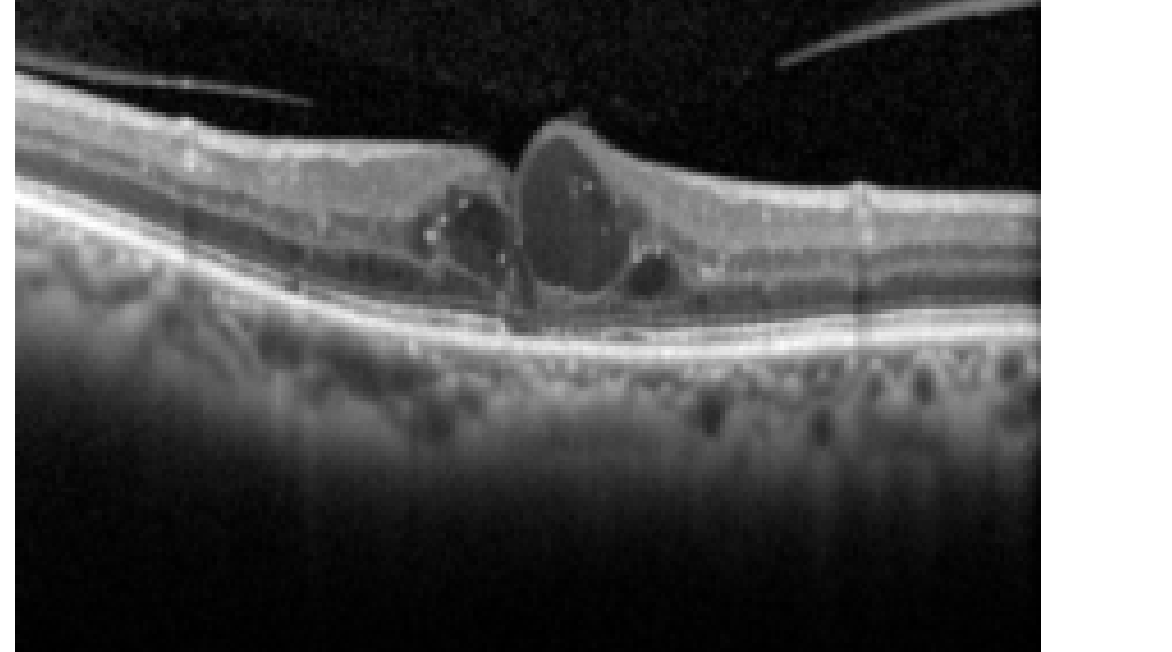}
    \hfill
    \includegraphics[width=0.48\linewidth,height=2cm]{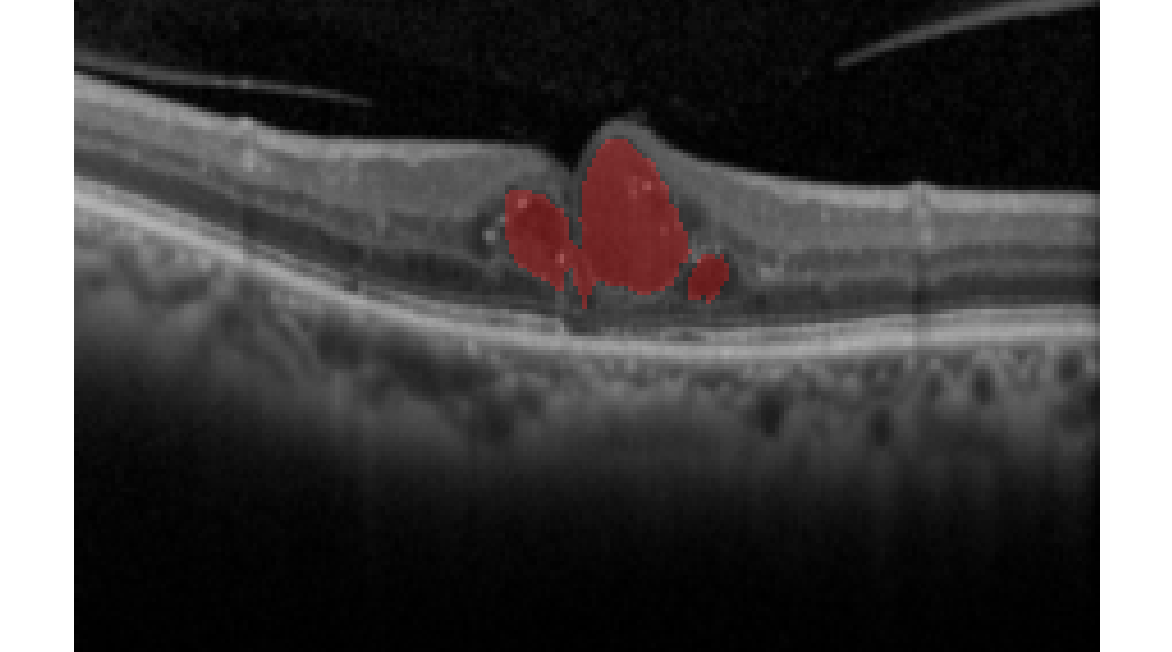}
    \caption{UMN}
    \label{fig:medical_umn}
\end{subfigure}
\hfill
\begin{subfigure}[t]{0.32\linewidth}
    \centering
    \includegraphics[width=0.48\linewidth,height=2cm]{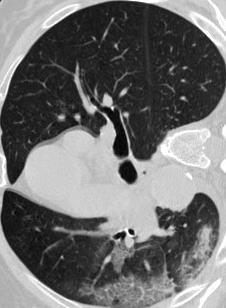}
    \hfill
    \includegraphics[width=0.48\linewidth,height=2cm]{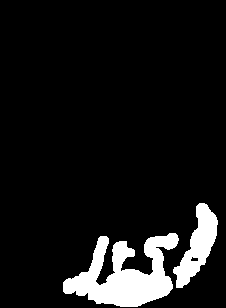}
    \caption{COVID-19}
    \label{fig:medical_covid}
\end{subfigure}

\caption{Example medical segmentation inputs and corresponding ground-truth masks for Duke, UMN, and COVID-19.}
\label{fig:medical-segmentation}
\end{figure}

\section{Background and Related Work}

In this section, we briefly introduce key concepts, including medical image segmentation and common segmentation models, \ac{dp} and \ac{dpsgd} in medical imaging, and improvements to \ac{dpsgd}, focusing on structure-aware methods and clipping strategies.

\subsection{Medical image segmentation}
Segmentation in medical imaging refers to the process of identifying and determining regions of interest, such as organs, tissues, and lesions \cite{ma2024segmenta}.
Deep learning methods achieve high accuracy by learning complex features, with UNet~\cite{ronneberger2015u} being a key milestone due to its encoder-decoder design with skip connections, enabling effective learning from limited data.

\acf{oct} is a modern imaging technique used to capture detailed tissue structure, such as retinal layers \cite{kafieh2013reviewa}. 
Segmentation of \ac{oct} images enables accurate retinal analysis and improve diagnosis \cite{pekala2019oct}. 
Similarly, \acf{ct} imaging is essential for analyzing lung diseases. 
For Lung CT COVID-19 Infection segmentation, many models are based on U-Net \cite{ronneberger2015u} and its variants. 

To enhance segmentation, topology-preserving methods enforce structural consistency (e.g., connectivity) \cite{hu2019topologypreservinga}, while shape-aware approaches incorporate global geometry, such as \ac{SDMs}, to guide smoother predictions \cite{tang2022shapeaware}.
Despite these advances, supervised learning methods require large amounts of annotated data, which are difficult to obtain and share due to privacy concerns. 
Moreover, sensitive training data can be inferred from trained models or gradients \cite{zhu2019advance}, highlighting the need for privacy-preserving approaches.


\subsection{Segmentation models}

State-of-the-art segmentation models are largely based on U-Net \cite{ronneberger2015u}. 
For OCT segmentation, we consider U-Net and two of its variants, LFUNet and NestedUNet, because they show high performance in this domain \cite{parsarad2025dpmorph}. 
For lung CT segmentation, we consider U-Net and NestedUNet, along with a specialized model for COVID-19 infection segmentation, Inf-Net \cite{fan2020infnet}.

\textbf{U-Net} consists of an encoder-decoder architecture with skip connections that improve detail preservation~\cite{ronneberger2015u}.
\textbf{LFUNet} combines a lightweight U-Net with full skip connections and deep supervision to improve training and mitigate vanishing gradients \cite{MA2021101988,li2022comprehensive}. 
It also incorporates dilated convolutions to enlarge the receptive field.
\textbf{Nested U-Net} (UNet++) redesigns skip connections with dense pathways and deep supervision, enabling better feature reuse and reducing the semantic gap between encoder and decoder \cite{zhou2018unet++}.

\textbf{Inf-Net} combines coarse-to-fine aggregation with attention to segment infection regions in CT, effectively handling high variation and low contrast \cite{fan2020infnet}. 
We use it as a strong baseline due to its progressive refinement of small and subtle lesions.

\subsection{\acf{dpsgd}}
\ac{dp} provides formal guarantees by limiting the influence of any single sample. 
A mechanism $\mathcal{R}$ satisfies $(\epsilon,\delta)$-DP if 
$
\Pr[\mathcal{R}(D)\in S] \leq e^{\epsilon}\Pr[\mathcal{R}(D')\in S] + \delta
$,
where smaller $\epsilon$ implies stronger privacy.
\acf{dpsgd}~\cite{abadi2016deep} enforces \ac{dp} by clipping per-sample gradients, adding Gaussian noise, and tracking privacy via a moments accountant, yielding an overall bound of $O(q\epsilon T,\delta)$.
While \ac{dpsgd} has been applied to medical imaging tasks \cite{ziller2021medical,mohammadi2026differential}, it often suffers from a considerable utility gap.

\subsection{\ac{dpsgd} in medical image segmentation}
Although \ac{dpsgd} has been applied in medical imaging, its use in segmentation remains limited. 
The deepee framework~\cite{ziller2021medical} shows near non-private performance on liver CT segmentation with UNet, but supports only fixed-norm clipping and lacks extensibility for alternative strategies.
More recent work \cite{ismail2025privacypreserving} evaluates several U-Net variants for dental CBCT segmentation under \ac{dpsgd}, showing only a slight performance drop with privacy.

Prior works report results under different privacy budgets, for example $\epsilon=1.5$ for CBCT lesion segmentation~\cite{ismail2025privacypreserving} 
and $\epsilon\approx2.7$ in the deepee framework \cite{ziller2021medical}.
However, these studies focus on binary segmentation tasks with large and very well-defined structures. 
In contrast, binary segmentation of small or irregular regions (e.g., OCT fluid or lung infection) and multi-class segmentation (e.g., retinal layers) remain underexplored and are more sensitive to \ac{dp} noise.

\ac{dpsgd} introduces a privacy--utility trade-off, often degrading accuracy and structural consistency. 
A promising direction for reducing the utility degradation is to integrate anatomically informed structural priors into neural architectures. 
While topology and shape constraints are well studied for improving segmentation \cite{hu2019topologypreservinga,tang2022shapeaware}, structural improvements for are underexplored. 
DP-Morph~\cite{parsarad2025dpmorph} integrates differentiable morphological operations for a lightweight local structural refinement while preserving $(\varepsilon,\delta)$-DP guarantees.
Morphological operators include \textit{erosion}, \textit{dilation}, \textit{opening}, and \textit{closing} \cite{GHOSH2024Multi}.
They apply a structuring element (a $k \times k$ kernel) to define a local neighborhood around each pixel.
Dilation computes the local maximum, $(I \oplus S)(x,y) = \max I(x+i, y+j)$, while erosion computes the local minimum, $(I \ominus S)(x,y) = \min I(x+i, y+j)$.
Opening (erosion followed by dilation) helps remove small artifacts, whereas closing (dilation followed by erosion) fills gaps and preserves structures.
DP-Morph is designed for flat clipping, where choosing the threshold $C$ is challenging. 
We extend it by evaluating \ac{NSGD}, \ac{Auto-S}, and \ac{PSAC}, and propose an adaptive approach with class-specific operations based on layer thickness.

\subsection{Clipping methods}
The standard \ac{dpsgd} algorithm \cite{abadi2016deep} uses flat clipping, which simplifies the privacy analysis by using a global clipping norm for all gradients. 
However, it has limitations: it truncates large gradients, leading to a loss of gradient magnitude, and requires careful tuning of the threshold $C$. 
Adaptive clipping approaches such as \ac{NSGD}~\cite{yang2022normalizeda}, \ac{Auto-S}~\cite{bu2023automaticc}, and \ac{PSAC}~\cite{xia2023differentiallya} address these limitations by dynamically adjusting the clipping to preserve gradient information under \ac{dp}.
These approaches are explained in the following and are compared in Table~\ref{tab:clipping-comparison}.

\textbf{\acf{NSGD}.}
\ac{NSGD} \cite{yang2022normalizeda} combines clipping with gradient normalization, scaling each gradient by its norm and a regularization term $r$. 
By enforcing equal gradients magnitudes, it couples the learning rate with clipping, reducing hyperparameter tuning.

\textbf{\acf{Auto-S}.}
\ac{Auto-S} \cite{bu2023automaticc} replaces hard clipping with a smooth normalization scheme that scales gradients by their norm plus a stability constant, $\gamma$, eliminating the need to tune a clipping threshold while preserving gradient magnitude information.
As proposed in \cite{bu2023automaticc}, we set $\gamma = 0.01$ as the default in our paper.

\textbf{\acf{PSAC}.}
\ac{PSAC}~\cite{xia2023differentiallya} extends \ac{dpsgd} by replacing hard clipping with a data-dependent scaling of per-sample gradient. 
Using a non-monotonic weight function based on the gradient norm reduces clipping bias and improves alignment with the true gradient while preserving \ac{dp}. 

\renewcommand{\arraystretch}{1.2}
\setlength{\tabcolsep}{6pt}

\begin{table*}[t]
\centering
\caption{Comparison of clipping strategies for \ac{dpsgd}.}
\label{tab:clipping-comparison}

\begin{tabularx}{\textwidth}{lccX
}
\toprule
\textbf{Method} & \textbf{Mechanism} & \textbf{Hard clip} & \textbf{Limitation} \\
\midrule

Flat~\cite{yousefpour2021opacus} & 
$\min(1, \frac{C}{\|g_i\|})$ & 
Yes &
Loss of magnitude information; requires tuning of $C$ \\

\ac{NSGD}~\cite{yang2022normalizeda} & 
$\frac{g_i}{r + \|g_i\|}$ & 
No &
Deviation from true batch gradient; depends on $r$ \\

\ac{Auto-S}~\cite{bu2023automaticc} & 
$\frac{g_i}{\|g_i\| + \gamma}$ & 
No &
Deviation from true batch gradient under small or imbalanced norms \\

\ac{PSAC}~\cite{xia2023differentiallya} &
$C \cdot \frac{g_i}{\|g_i\| + \frac{r}{\|g_i\|+r}}$ &
No &
Additional design complexity due to norm-dependent scaling \\

\bottomrule
\end{tabularx}
\end{table*}


\section{Methodology}\label{sec:methodology}

We evaluate \ac{oct} and \ac{ct} segmentation under private and non-private settings, comparing multiple clipping strategies with and without morphological refinement. 
For multi-class tasks (e.g., Duke), we propose an adaptive morphology based on class thickness (e.g., retinal layers) to select operations dynamically.

\subsection{Integration of clipping strategies into segmentation models}

Standard architectures, such as UNet, UNet++, are not directly compatible with \ac{dpsgd}, mainly due to Batch Normalization (BatchNorm), which violates per-sample gradient requirements. 
Thus, it is replaced with GroupNorm or LayerNorm, with GroupNorm being more suitable because it better preserves channel-wise feature structure in convolutional models. 
After ensuring architectural compatibility with \ac{dpsgd}, we integrate clipping strategies into the training pipeline using modified version of Opacus.

In standard \ac{dpsgd}, per-sample gradients are clipped using a fixed threshold $C$.
In our implementation, we replace this operation with alternative clipping rules, including adaptive (\ac{Auto-S} and \ac{NSGD}) and per-sample schemes (\ac{PSAC}), which dynamically scale gradients based on their norms.
These modifications are applied prior to gradient aggregation and noise addition, ensuring compatibility with the \ac{dpsgd} pipeline.

\subsection{Adaptive DP-Morph for multi-class segmentation}
We propose a class-adaptive morphology policy for multi-class segmentation, focusing on retinal layers due to their thin, structured, and noise-sensitive nature. 
Figure \ref{fig:medical_duke} shows an example from the Duke dataset, where different retinal layers are visualized in distinct colors.
The segmentation consists of 9 classes: background (0), retinal layers (1-7), and fluid (8), ordered from top to bottom.
In adaptive DP-Morph, instead of uniform operations, we estimate class-wise thickness from predictions and assign morphology accordingly. 
The method includes: (i) thickness estimation, (ii) batch-level thresholding, and (iii) class-wise operation assignment with refinement.


\textbf{Thickness estimation.}
We estimate layer thickness from model predictions rather than ground truth annotations for multiple reasons.
First, incorporating such information would turn the morphological module into a label-driven component and introduce an unrealistic training advantage, as this information is not available at inference time.
Second, prediction-based estimation allows the operations to adapt to the current state of the model.

Given network logits $z \in \mathbb{R}^{B \times C \times H \times W}$, we first obtain the predicted segmentation map:
$
\hat{y}_{b,y,x} = \arg\max_{c} z_{b,c,y,x}.
$
For each image $b$ and class $c$, we estimate the thickness as the average number of pixels assigned to class $c$ along the vertical axis, averaged over all columns:
$
t_{b,c} = \frac{1}{W} \sum_{x=1}^{W} \sum_{y=1}^{H} \mathbf{1}\big[\hat{y}_{b,y,x} = c\big].
$

This results in a thickness matrix: $T \in \mathbb{R}^{B \times C},$
where $T_{b,c}$ represents the average thickness of class $c$ in image $b$.
In OCT, retinal layers are horizontally organized, making per-column pixel counts a natural estimate of thickness.

\textbf{Batch-level threshold estimation.}
We compute thickness thresholds at the batch level rather than using global statistics. 
Global thresholds, even if estimated from the latest model state (e.g., at the end of the previous epoch), may not accurately reflect the current prediction distribution, particularly under \ac{dp} training where noise introduces additional variability.
Moreover, estimating thresholds at the current epoch level requires a two-pass procedure: one pass to collect predictions and compute thickness statistics, and a second pass to apply morphological operations. This increases computational cost and breaks the standard training pipeline, where predictions and transformations are applied jointly within each iteration.
In contrast, per-batch thresholds are computed directly from the same predictions to which morphology is applied, ensuring consistency with the current model state. Although batch composition influences the thresholds, they capture relative thickness within the current prediction distribution.

Let $\mathcal{R}$ denote the set of retinal layer classes (e.g., $\mathcal{R} = \{1,\dots,7\}$). 
We collect all thickness values for retinal classes across the batch and compute lower and upper quantile thresholds:
\begin{equation}
T_{\text{thin}} = Q_{q_{\text{low}}} \big( \{ t_{b,c} \mid c \in \mathcal{R} \} \big),
\qquad
T_{\text{thick}} = Q_{q_{\text{high}}} \big( \{ t_{b,c} \mid c \in \mathcal{R} \} \big),
\end{equation}
where $Q_q(\cdot)$ denotes the $q$-th quantile. 
We compare multiple thresholding strategies (Table~\ref{tab:quantile_comparison_nestedunet_all_policies} in the Appendix) and adopt $(q_{\text{low}}=20, q_{\text{high}}=85)$ due to its superior performance.
We then compute the batch-averaged thickness per class:
\begin{equation}
\bar{t}_c = \frac{1}{B} \sum_{b=1}^{B} t_{b,c}.
\end{equation}

\textbf{Operation assignment policy.}
We define three thickness-aware morphology policies that assign operations based on class thickness. 
All policies apply closing to thin classes and a combined operation (opening followed by closing) to intermediate classes.
They differ in the treatment of thick layers: v1 applies opening (noise removal), v2 applies both operations (balanced refinement), and v3 applies closing (structure preservation), enabling controlled exploration of the trade-off between denoising and structural fidelity.
For thin layers, we consistently apply closing, as these structures are highly susceptible to fragmentation and discontinuities. 
While noise may still be present, operations involving erosion (such as opening) risk removing the layer entirely.
Therefore, closing provides a safer choice by preserving connectivity and bridging gaps.

\section{Implementation}
\label{sec:implementations}

This section outlines the implementation details of our framework, including the datasets, evaluation metrics, and experimental setup.

\subsection{Datasets}
\label{sec:datasets}
\textbf{OCT Datasets}.
We evaluate \ac{oct} segmentation on the Duke and UMN datasets. 
The UMN DME dataset contains OCT scans from 29 subjects with expert annotations \cite{rashno2017fully}, while the Duke dataset includes 110 annotated B-scans with eight retinal layers \cite{chiu2015kernel}. 
Both use a 60/20/20 train/validation/test split.
\textbf{Lung CT Images}.
For lung infection segmentation, we use the COVID-19 CT dataset~\cite{medseg_covid19_dataset}, consisting of 100 annotated CT images from Italian Society of Medical and Interventional Radiology (SIRM). 
This dataset is also used to train Inf-Net~\cite{fan2020infnet}.
Figure \ref{fig:medical-segmentation} shows a sample from each of our datasets.

\subsection{Evaluation Metrics}\label{subsec:metrics}
We use Dice and MAE as common evaluation metrics across all datasets to enable consistent comparison, capturing pixel-wise accuracy.
For COVID-19, we also report structure-aware metrics (\ac{se}, \ac{ee}) \cite{fan2020infnet}, while for OCT we use \ac{HD95} to assess boundary alignment of thin layers \cite{taha2015metrics}.

\noindent\textbf{Metrics for segmentation performance.}
The \textit{Dice Coefficient} measures overlap between predictions and ground truth, with 1 indicating perfect agreement and 0 no overlap, and is defined as \cite{zou2004statistical}: 
$
\text{Dice} = \frac{2 \times |\hat{y} \cap y|}{|\hat{y}| + |y|}
$,
where $y$ denotes the ground truth mask and $\hat{y}$ the predicted mask. 
%
The \textit{Mean Absolute Error} (MAE) measures the average absolute difference between predictions and ground truth across all classes \cite{viedma2022oct}.

\noindent\textbf{Task specific metrics: OCT segmentation.}
The \textit{95th Percentile Hausdorff Distance} (HD95) measures boundary mismatch using the 95th percentile of bidirectional distances between predicted and ground-truth boundaries \cite{taha2015metrics}. 
Lower values indicate better alignment \cite{taha2015metrics}; we compute it per class and average over classes (excluding background).
If $\partial \hat{y}$ and $\partial y$ denote the sets of boundary points of the predicted and ground-truth segmentations, then we define the collection of bidirectional nearest-neighbor distances as:
\begin{equation}
\mathcal{HD} = \left\{ \min_{q \in \partial y} d(p,q) \mid p \in \partial \hat{y} \right\}
\cup
\left\{ \min_{p \in \partial \hat{y}} d(p,q) \mid q \in \partial y \right\},
\end{equation}
where $d(\cdot,\cdot)$ denotes the Euclidean distance.

\noindent\textbf{Task specific metrics: Lung CT segmentation.}
The \textit{\acf{se}}~\cite{fan2020infnet} evaluates regional and object-level structural similarity between $\hat{y}$ and $y$: $S_{\alpha} = (1 - \alpha) \cdot S_o(\hat{y}, y) + \alpha \cdot S_r(\hat{y}, y)$,
This metric captures global shape and spatial structure, aligning better with human perception than pixel-wise measures.
The \textit{\acf{ee}}~\cite{fan2020infnet} evaluates local and global consistency by combining pixel-level and image-level statistics between $\hat{y}$ and $y$:
\begin{equation}
   E_{\phi} = \frac{1}{w \times h} \sum_{i=1}^{w} \sum_{j=1}^{h} \phi(\hat{y}(i,j), y(i,j)),
\end{equation}
where $w$ and $h$ denote the image width and height, respectively, and $\phi$ is the enhanced alignment function. 
Following standard practice, we compute $E_{\phi}$ over thresholds 0-255 and report the mean, where higher values indicate better alignment between $\hat{y}$ and $y$.

\subsection{Experimental design}
\label{sec:parameters}


We implemented all experiments in PyTorch with Opacus. 
We extend Opacus to support adaptive methods (\ac{Auto-S}, \ac{NSGD}), and per-sample clipping (\ac{PSAC}).

Experiments are run on an NVIDIA RTX 4090 GPU. 
For the Duke dataset, evaluation is restricted to retinal layers, excluding background and fluid to avoid bias the evaluation of layer segmentation performance. 
We also compare all methods with and without DP-Morph to study its interaction with clipping strategies.
The implementation of all components, including the extended Opacus framework and DP-Morph, is publicly available to ensure reproducibility~\cite{Parsarad2026git}.

In many medical imaging contexts, relatively large privacy budgets (e.g., $\epsilon \approx 10^{6}$) can still effectively mitigate reconstruction attacks \cite{ziller2024reconciling}. 
To be consistent with prior work~\cite{parsarad2025dpmorph,ziller2024reconciling}, for our segmentation tasks, we mainly consider a privacy budget of $\varepsilon = 200$.
We evaluate morphological operations (open, close, both) and kernel sizes ($k \in {3,5}$), reporting the best-performing variant. All results are averaged over three runs for consistency.

\textbf{Ablation studies for flat clipping.}
Flat clipping performance depends on the learning rate and clipping threshold. 
We perform an ablation over both, averaging four runs per setting to evaluate Dice, with thresholds guided by \cite{bu2023automaticc}. 
Figure~\ref{fig:ablation} shows results for UNet on Duke and UMN (batch size 8), with similar trends observed across models.

\begingroup
\setlength{\textfloatsep}{8pt}
\begin{figure}[!t]
  \centering

  \begin{subfigure}{0.48\linewidth}
    \centering
    \includegraphics[width=\linewidth]{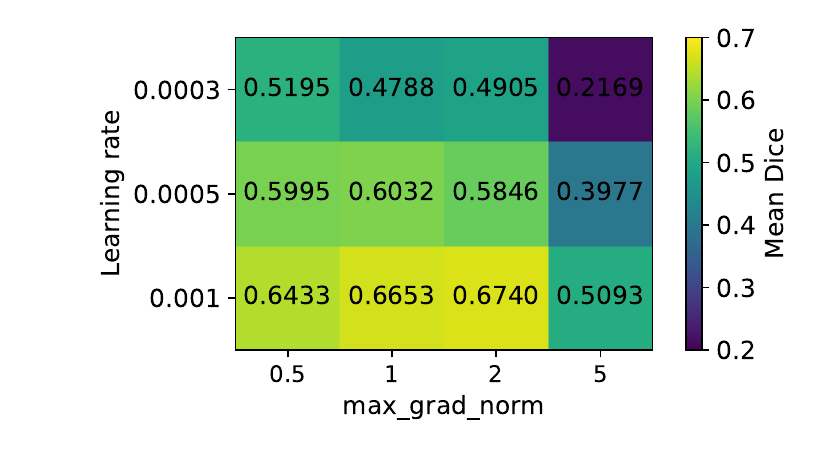}
    \caption{Duke}
    \label{fig:duke_abl}
  \end{subfigure}
  \hfill
  \begin{subfigure}{0.48\linewidth}
    \centering
    \includegraphics[width=\linewidth]{ 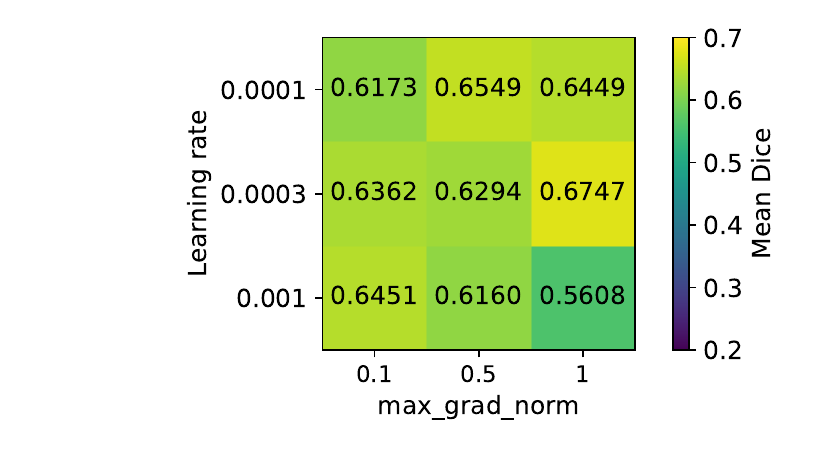}
    \caption{UMN}
    \label{fig:umn_abl}
  \end{subfigure}
  \caption{Heatmaps of UNet performance under flat clipping for different learning rates and clipping thresholds (batch size = 8) on Duke and UMN.}
  \label{fig:ablation}
\end{figure}
Consistent with \cite{bu2023automaticc}, the learning rate has a stronger effect than the clipping threshold. 
For Duke, we use $0.001$ with $C=1$ (similar results for $C \in [0.5,1,2]$), while for UMN, $0.0003$ with $C=1$ performs best. 
For the Lung CT dataset, we follow the training configuration of \cite{fan2020infnet} as a baseline for learning rate and batch size. 
Within this setting, we perform additional tuning for \ac{dpsgd} and find that a clipping threshold of $C = 1.5$ shows the best empirical performance.
\section{Results}
We report segmentation performance on OCT and lung CT across clipping strategies, with and without morphology, and analyze gradient differences, runtime, and privacy via a global loss attack. 
We also evaluate the proposed adaptive approach on the Duke multi-layer segmentation task.

\subsection{Impact of clipping strategies on model performance} 
%
In this section, we present the impact of clipping strategies with DP-Morph, across Duke, UMN, and Covid-19 (Lung CT) datasets.

\textbf{Duke dataset.}
We evaluate UNet++, UNet, and LFUNet. Figure~\ref{fig:metrics_duke} shows results for UNet and UNet++, while LFUNet results are in Table~\ref{tab:combined_models_best_duke} (Appendix).
Figure~\ref{fig:metrics_duke} shows that both UNet and UNet++ consistently benefit from morphological refinement in the non-private setting. 
For UNet, morphology yields an improvement of approximately 0.6\% in Dice, while UNet++ shows a comparable gain of around 0.5\%.
Similarly, boundary accuracy improves substantially, with \ac{HD95} reduced by roughly 17\% for UNet and 19\% for UNet++. 

Under \ac{dpsgd}, UNet++ is more robust to noise, achieving higher Dice across most clipping strategies. \ac{Auto-S} and Flat perform best (Dice $\sim$0.71), while \ac{NSGD} shows the weakest results, particularly for UNet without morphology. 
Morphological refinement mitigates this degradation, improving Dice by $\sim$30\% and reducing \ac{HD95} by $\sim$17\%.
Figure \ref{fig:qualitative_results_combined} shows a visual example of the Duke dataset under clipping strategies for the DP-Morph version of U-Net++. 

\begin{figure}[!t]
  \centering

  \begin{subfigure}[t]{\textwidth}
    \centering
    \includegraphics[width=0.75\textwidth]{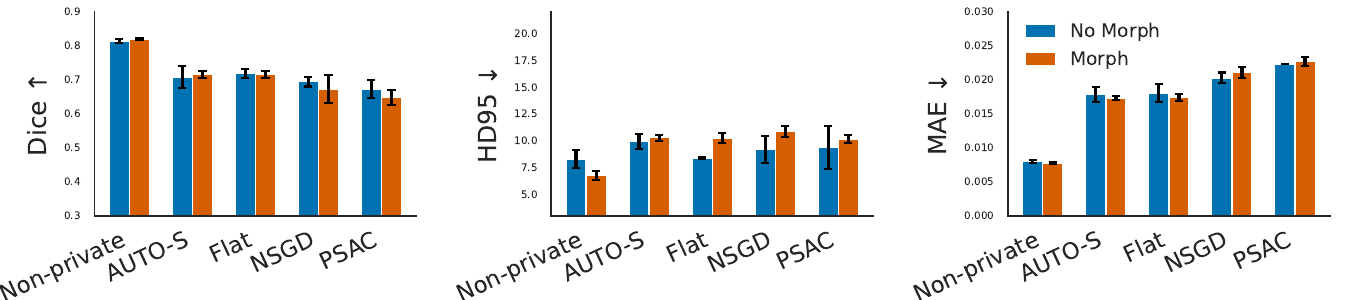}
    \caption{UNet++}
    \label{fig:umn_nestedunet}
  \end{subfigure}

  \vspace{0.5em}

  \begin{subfigure}[t]{\textwidth}
    \centering
    \includegraphics[width=0.75\textwidth]{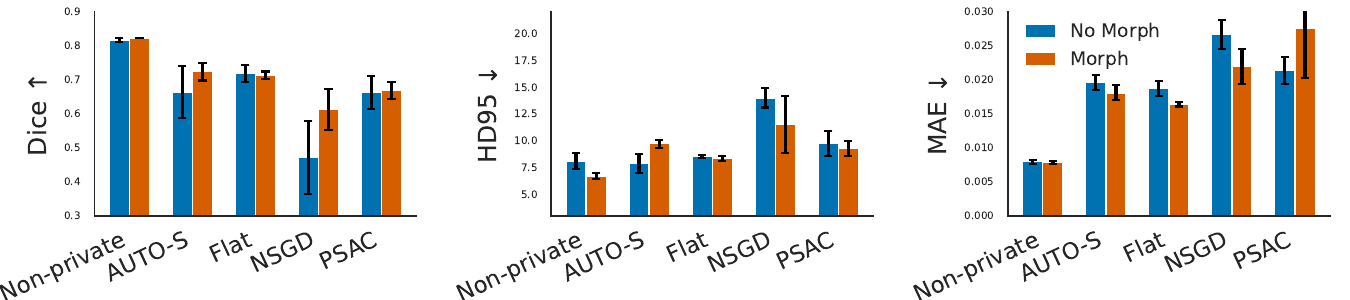}
    \caption{UNet}
    \label{fig:umn_unet}
  \end{subfigure}

  \caption{Comparison of clipping strategies on the Duke dataset for UNet++ and UNet.}
  \label{fig:metrics_duke}
\end{figure}

\textbf{UMN dataset.}
Figure~\ref{fig:metrics_umn} shows that both architectures consistently improve in Dice when morphological refinement is applied. 
In the non-private setting, UNet++ achieves the best performance with morphology, corresponding to a relative improvement of over 6\% compared to its no-morph counterpart.

Under \ac{dpsgd}, \ac{Auto-S} and Flat achieve the best performance, with UNet++ reaching Dice $\sim$0.71--0.73 when combined with morphology, while \ac{NSGD} remains competitive.
In contrast, \ac{PSAC} consistently performs worst for both architectures, with a severe degradation of $\sim$40--50\% compared to other strategies.

Morphological refinement consistently improves Dice across all clipping strategies, partially mitigating performance drops, particularly for weaker methods such as \ac{PSAC} (improving Dice by $\sim$10--25\%). 
However, despite these gains, \ac{PSAC} remains the least effective overall.
In the non-private setting, morphology also improves Dice (up to $\sim$6\% for UNet++), although \ac{HD95} differences remain small.
An UMN segmentation visual results with different clipping strategies for unet++  is shown in Figure \ref{fig:qualitative_results_combined}.

\begin{figure}[!t]
  \centering

  \begin{subfigure}[t]{\textwidth}
    \centering
    \includegraphics[width=0.75\textwidth]{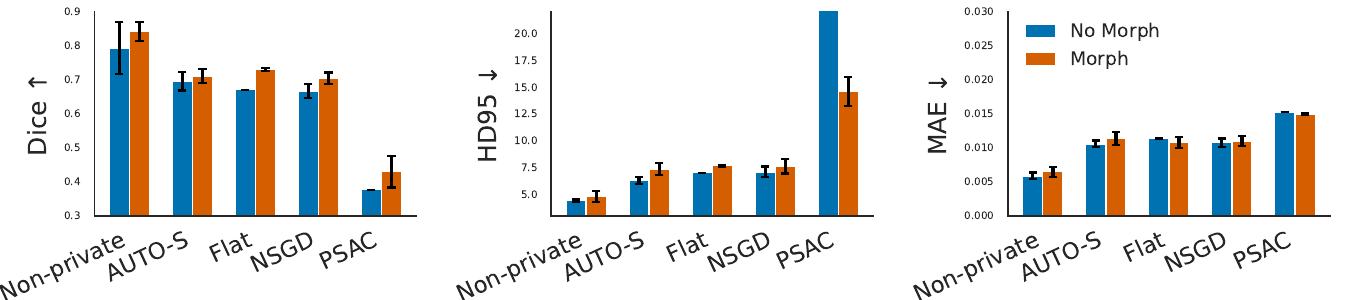}
    \caption{UNet++}
    \label{fig:umn_nestedunet_clip_strategies}
  \end{subfigure}

  \vspace{0.5em}

  \begin{subfigure}[t]{\textwidth}
    \centering
    \includegraphics[width=0.75\textwidth]{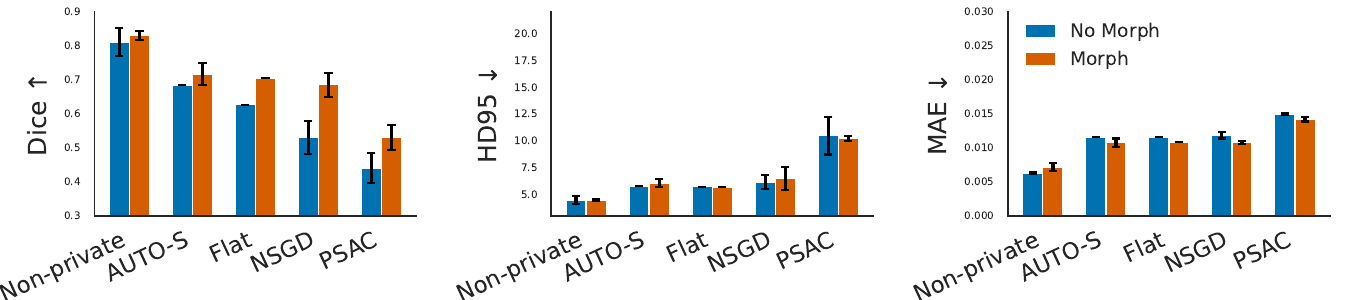}
    \caption{UNet}
    \label{fig:umn_unet_clip_strategies}
  \end{subfigure}

  \caption{Comparison of clipping strategies on the UMN dataset for UNet++ and UNet.}
  \label{fig:metrics_umn}
\end{figure}

\textbf{COVID-19 Lung CT dataset.}
Figure \ref{fig:metrics_lunct} shows the results on the COVID-19 Lung CT dataset.
We focus on UNet++ and Inf-Net, as UNet++ performs best under \ac{dpsgd}, while Inf-Net achieves the strongest non-private performance.
U-Net is omitted for brevity due to consistently lower performance.

UNet++ shows greater robustness under \ac{dpsgd} than Inf-Net, achieving higher Dice in private settings ($\sim$0.50 vs.\ $\sim$0.45).
While Inf-Net performs best in the non-private regime, it shows a larger drop under privacy constraints ($\sim$0.22 vs.\ $\sim$0.11 for UNet++).

Morphological refinement improves performance in the non-private setting (Dice gains of $\sim$1--2\%) and maintains comparable or slightly improved results under \ac{dpsgd}.
Among clipping strategies, \ac{Auto-S} and \ac{PSAC} better preserve performance with morphology, while \ac{NSGD} shows greater degradation. 
These trends are consistent with the visual results (Figure~\ref{fig:qualitative_results_combined}).

\begin{figure}[!t]
  \centering
\begin{subfigure}[t]{\textwidth}
    \centering
    \includegraphics[width=1\linewidth]{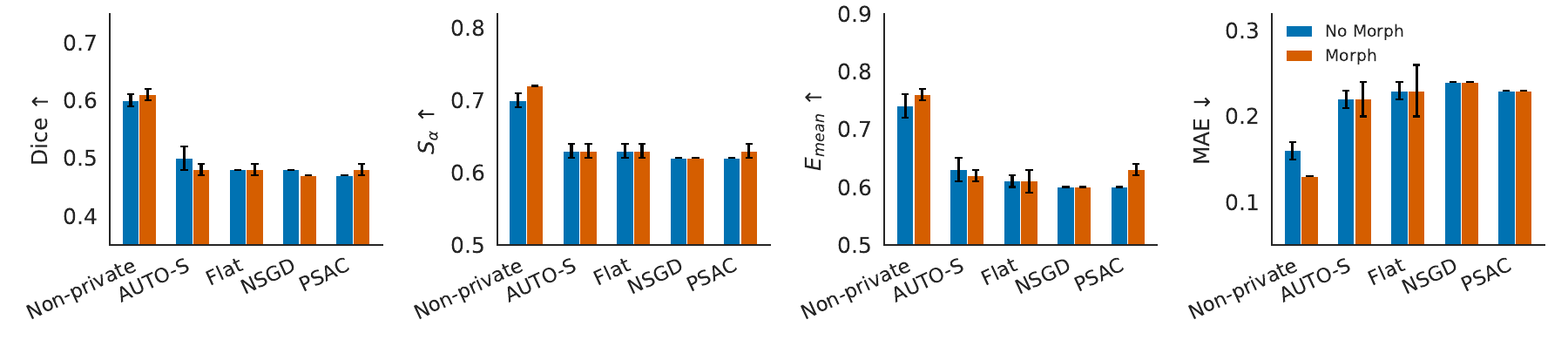}
    \caption{UNet++}
    \label{nestedunet:lungct}
  \end{subfigure}

  \begin{subfigure}[t]{1\textwidth}
    \centering
    \includegraphics[width=1\linewidth]{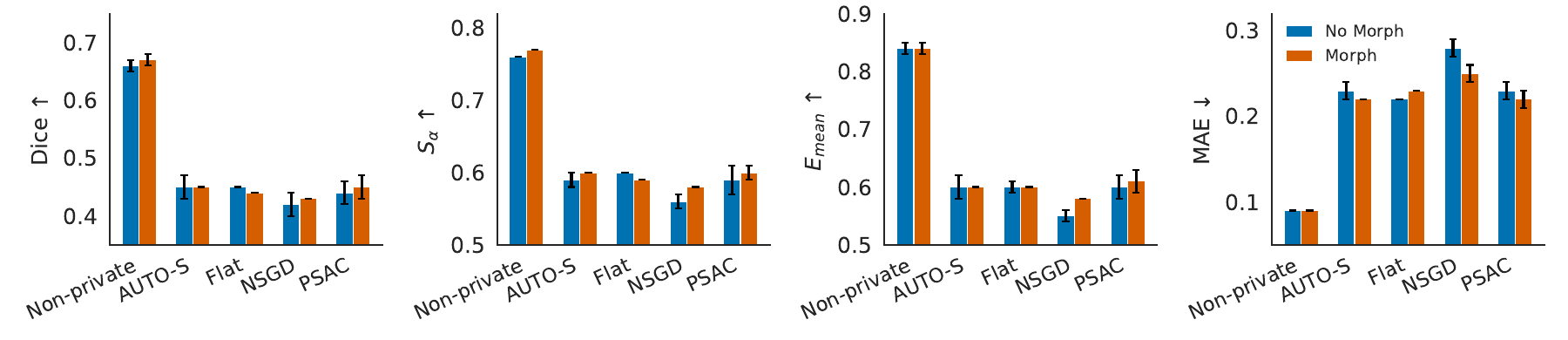}
    \caption{Inf-Net}
    \label{Infnet:lungct}
  \end{subfigure}
  \caption{Comparison of clipping strategies on the Covid-19 dataset for
UNet++ and Inf-Net.}
  \label{fig:metrics_lunct}

\end{figure}

\begin{figure}[!t]
\centering
\scriptsize
\setlength{\tabcolsep}{2pt}

\resizebox{0.95\textwidth}{!}{
\begin{tabular}{@{}c c c c c c c@{}}
\textbf{Input} &
\textbf{Ground truth} &
\shortstack{\textbf{InfNet}\\\textbf{Non-Private}} &
\textbf{Auto-S} &
\textbf{Flat} &
\textbf{NSGD} &
\textbf{PSAC} \\[0.2em]

\includegraphics[width=1.8cm,height=1.8cm,keepaspectratio]{Figures/lungct_segmentation/unet/unet_sample6_input.jpg} &
\includegraphics[width=1.8cm,height=1.8cm,keepaspectratio]{Figures/lungct_segmentation/unet/unet_sample6_gt.png} &
\includegraphics[width=1.8cm,height=1.8cm,keepaspectratio]{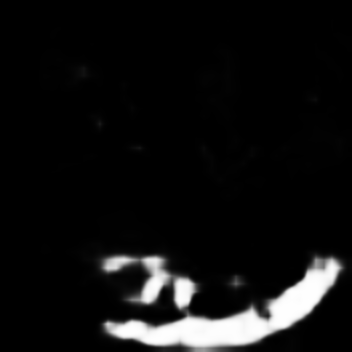} &
\includegraphics[width=1.8cm,height=1.8cm,keepaspectratio]{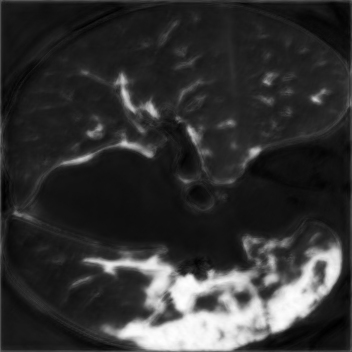} &
\includegraphics[width=1.8cm,height=1.8cm,keepaspectratio]{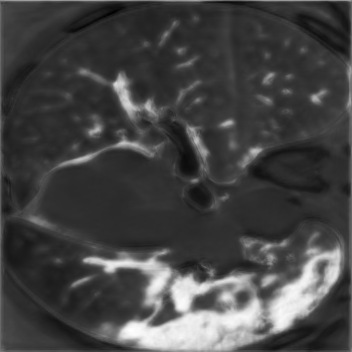} &
\includegraphics[width=1.8cm,height=1.8cm,keepaspectratio]{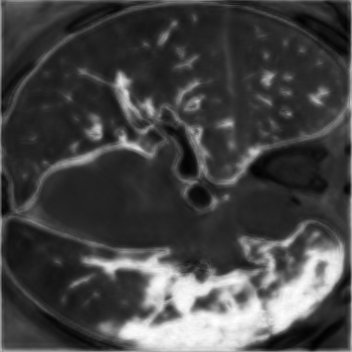} &
\includegraphics[width=1.8cm,height=1.8cm,keepaspectratio]{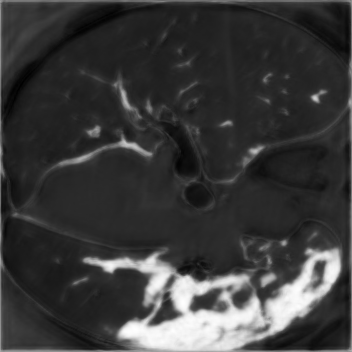} \\
\end{tabular}
}

\vspace{0.4em}

\resizebox{0.95\textwidth}{!}{
\begin{tabular}{@{}c c c c c c c@{}}
\textbf{Input} &
\textbf{Ground truth} &
\shortstack{\textbf{Non-}\textbf{Private}} &
\textbf{Auto-S} &
\textbf{Flat} &
\textbf{NSGD} &
\textbf{PSAC} \\[0.2em]

\includegraphics[width=1.8cm,height=1.8cm,keepaspectratio]{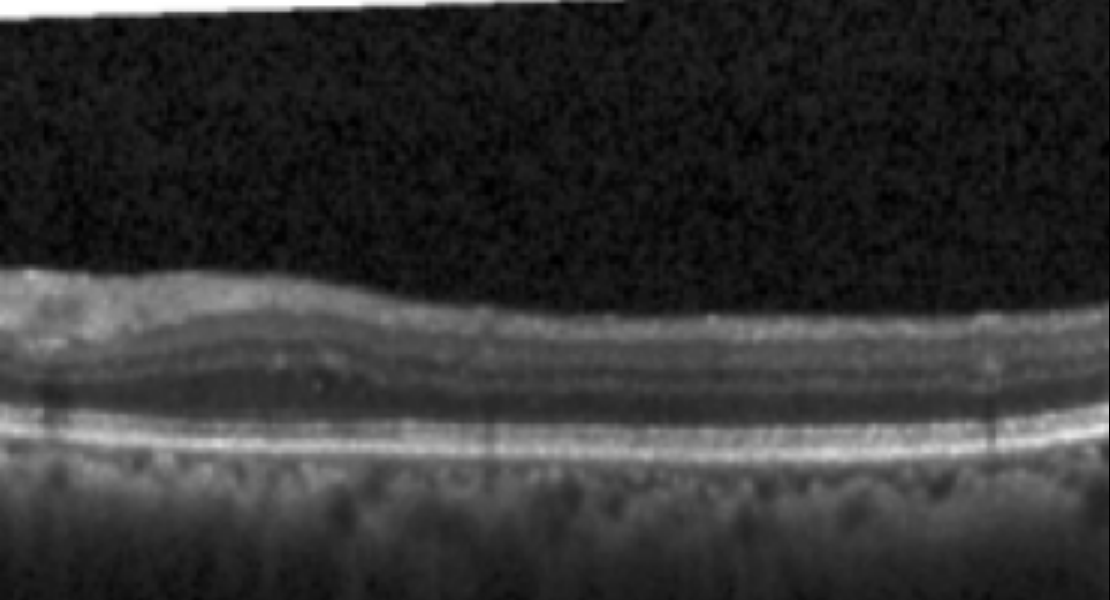} &
\includegraphics[width=1.8cm,height=1.8cm,keepaspectratio]{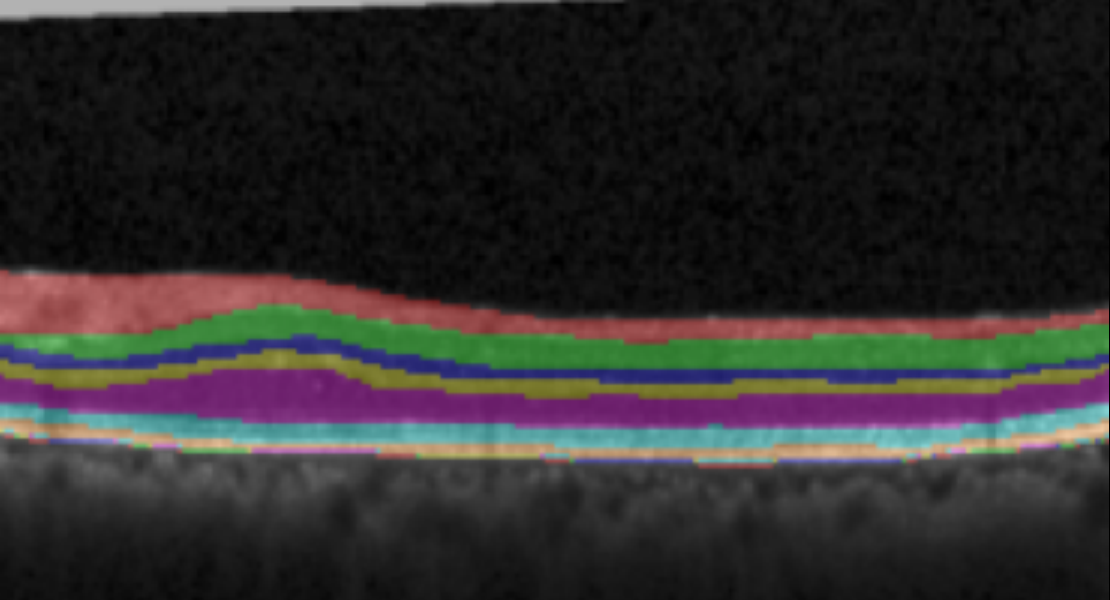} &
\includegraphics[width=1.8cm,height=1.8cm,keepaspectratio]{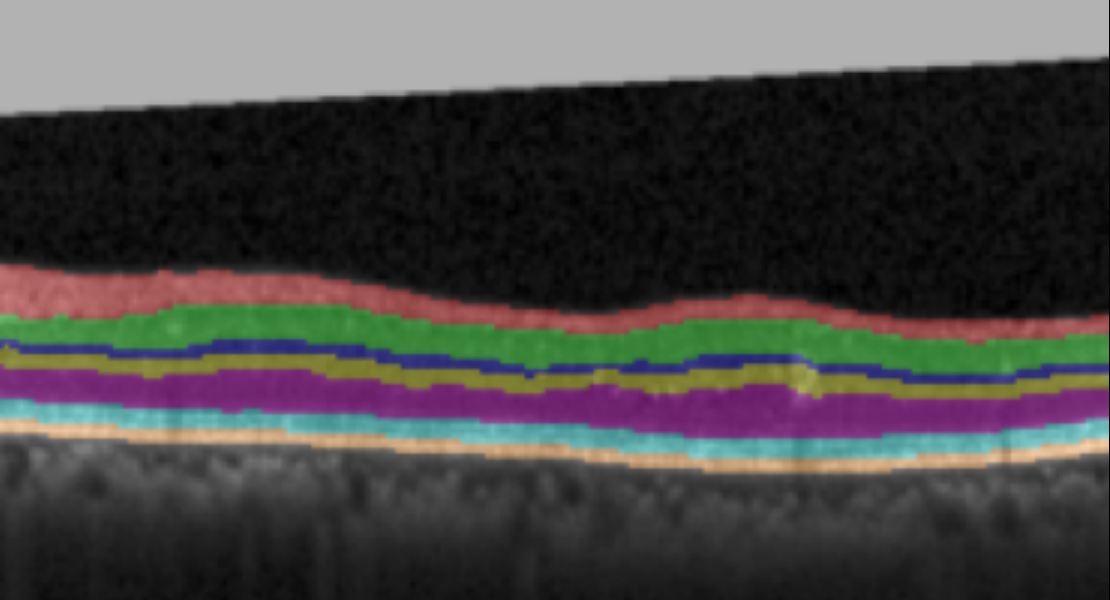} &
\includegraphics[width=1.8cm,height=1.8cm,keepaspectratio]{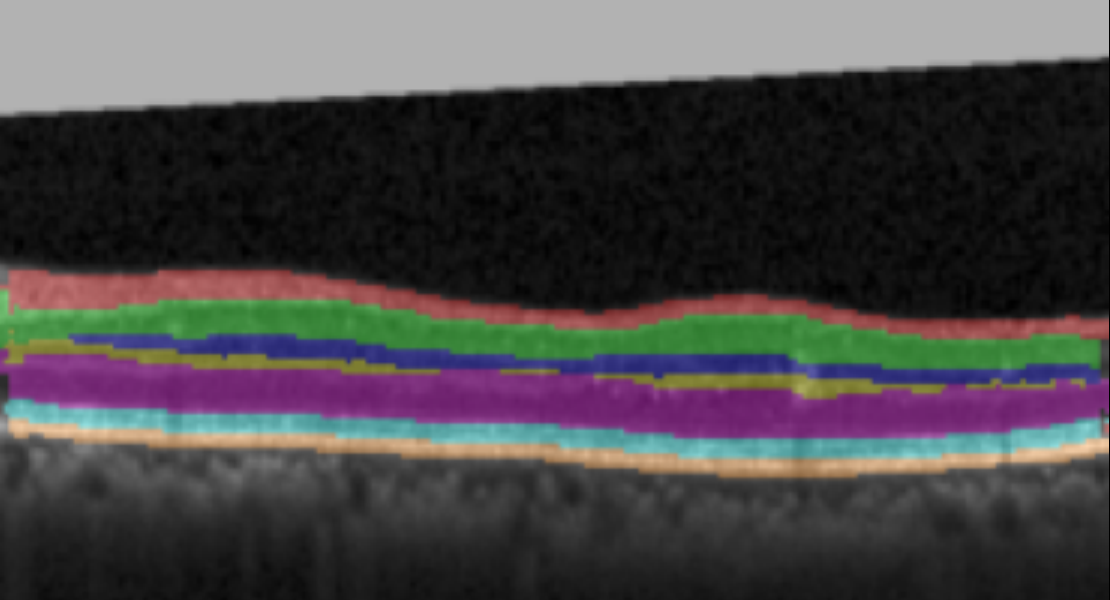} &
\includegraphics[width=1.8cm,height=1.8cm,keepaspectratio]{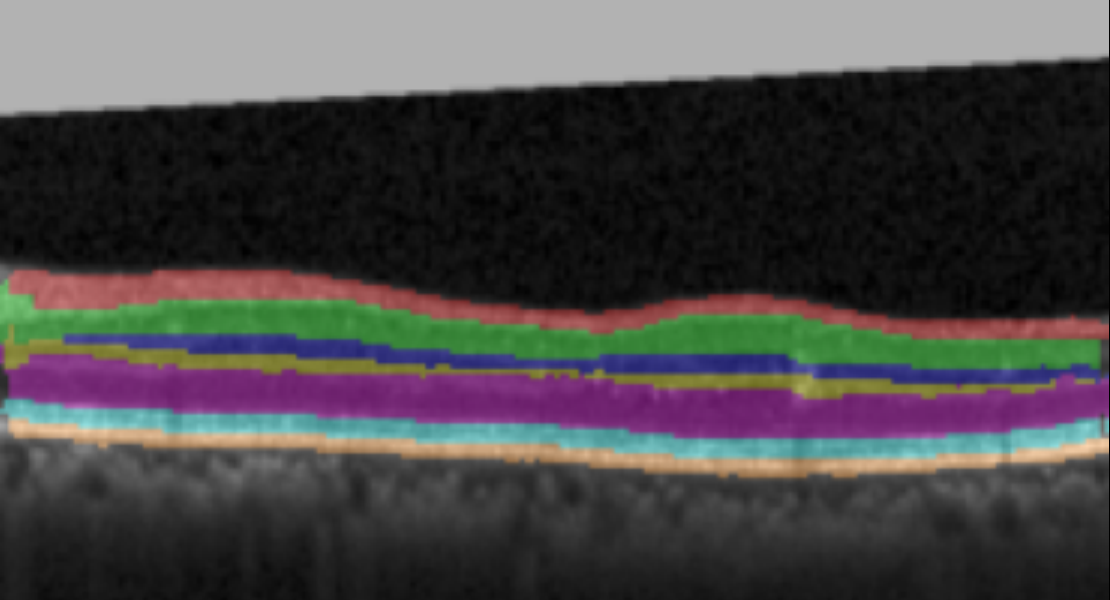} &
\includegraphics[width=1.8cm,height=1.8cm,keepaspectratio]{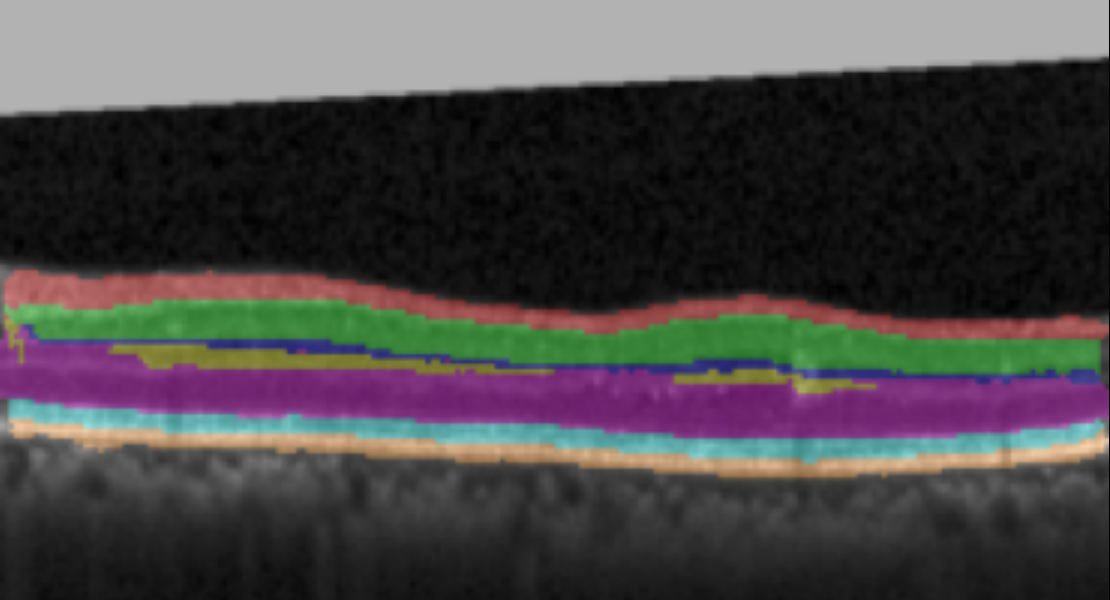} &
\includegraphics[width=1.8cm,height=1.8cm,keepaspectratio]{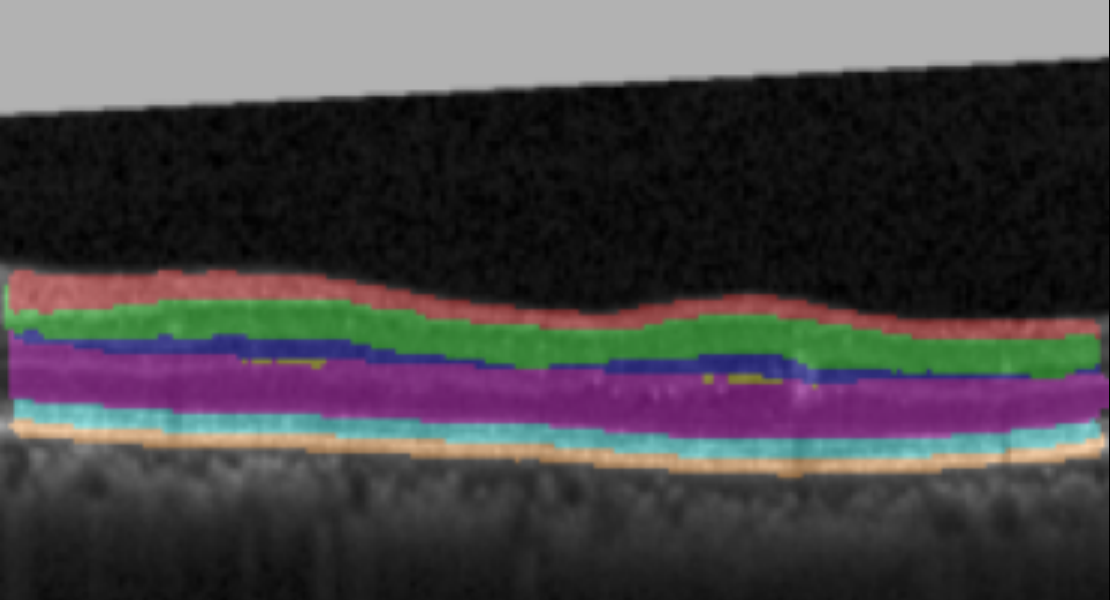} \\[0.2em]

\includegraphics[width=1.8cm,height=1.8cm,keepaspectratio]{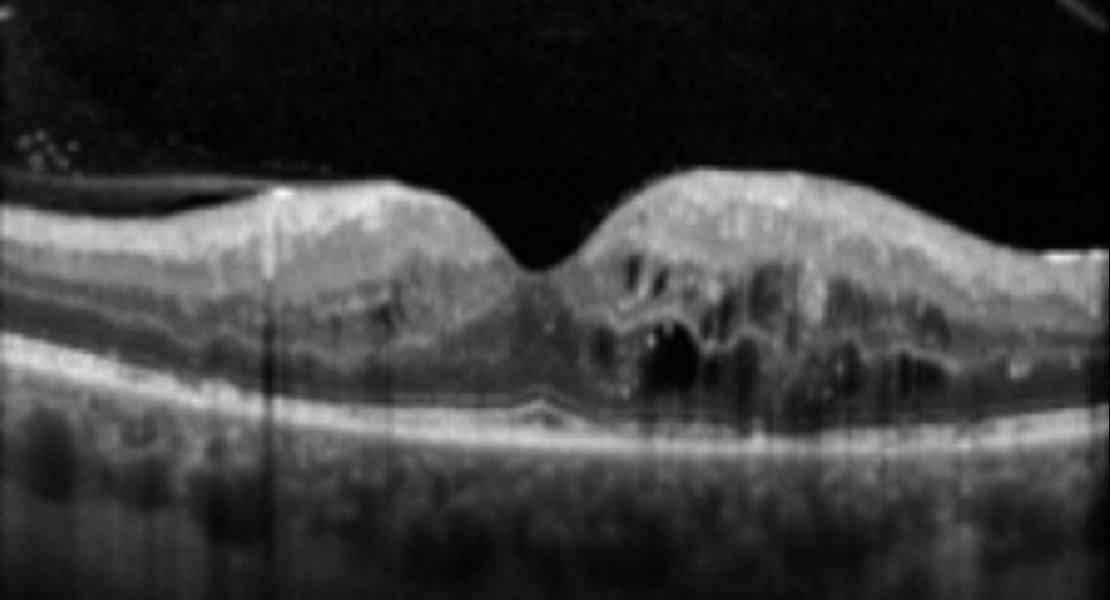} &
\includegraphics[width=1.8cm,height=1.8cm,keepaspectratio]{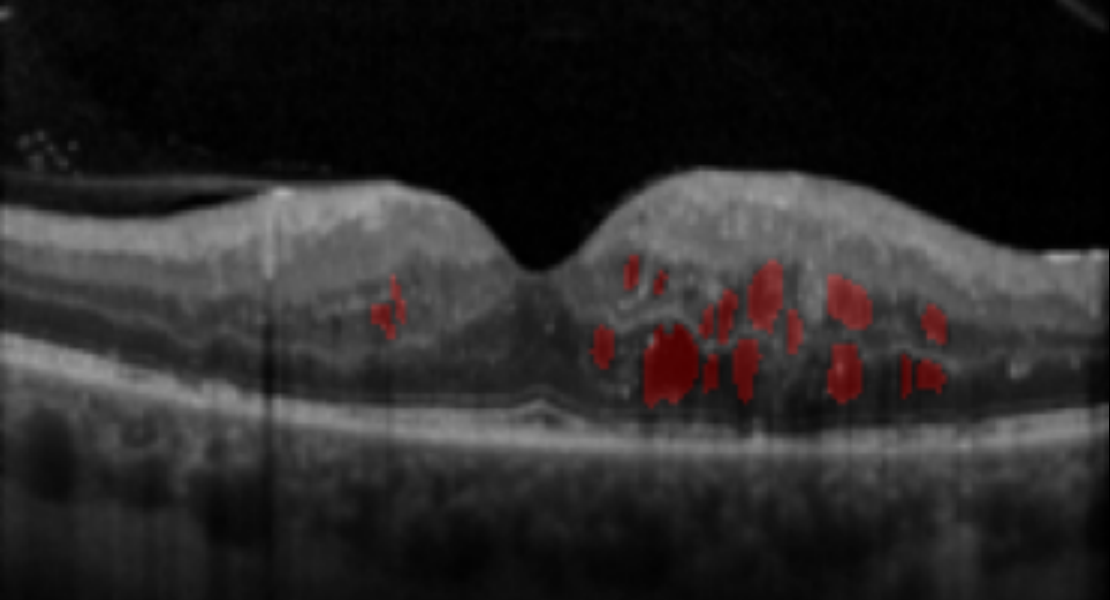} &
\includegraphics[width=1.8cm,height=1.8cm,keepaspectratio]{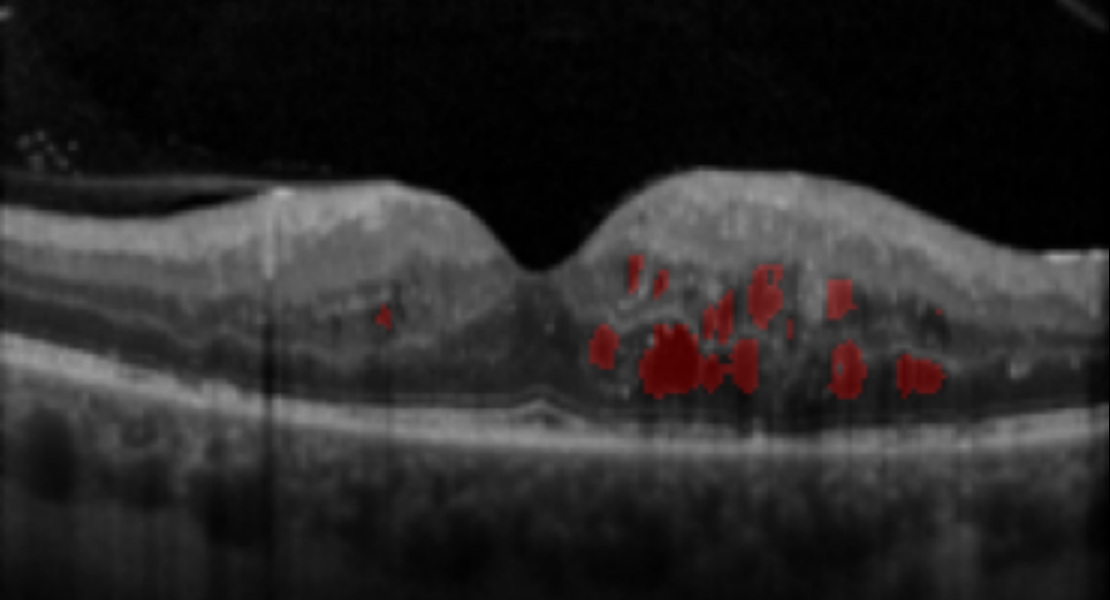} &
\includegraphics[width=1.8cm,height=1.8cm,keepaspectratio]{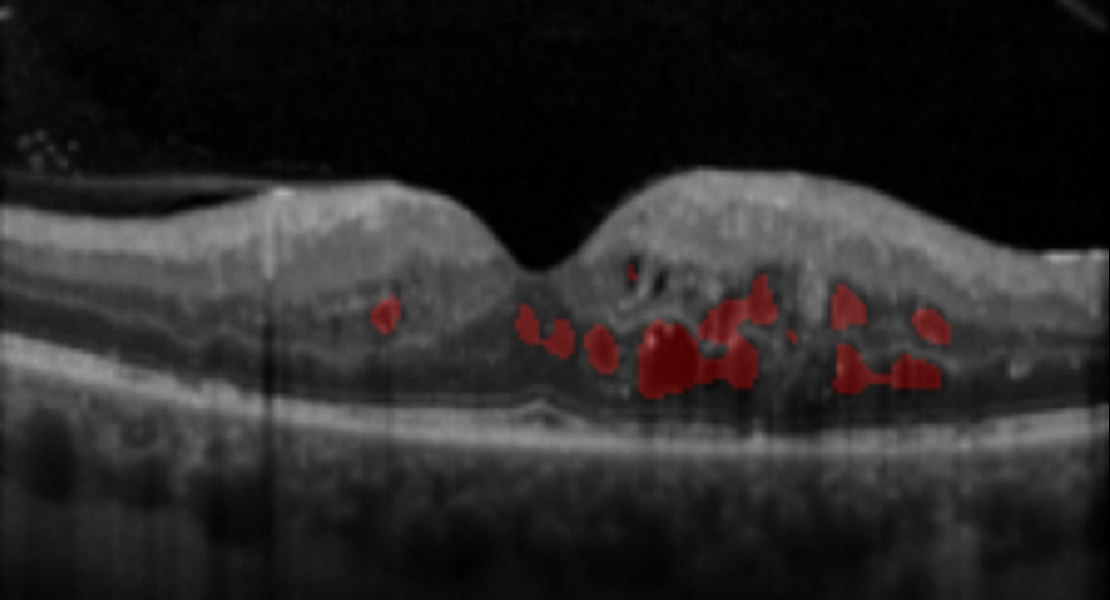} &
\includegraphics[width=1.8cm,height=1.8cm,keepaspectratio]{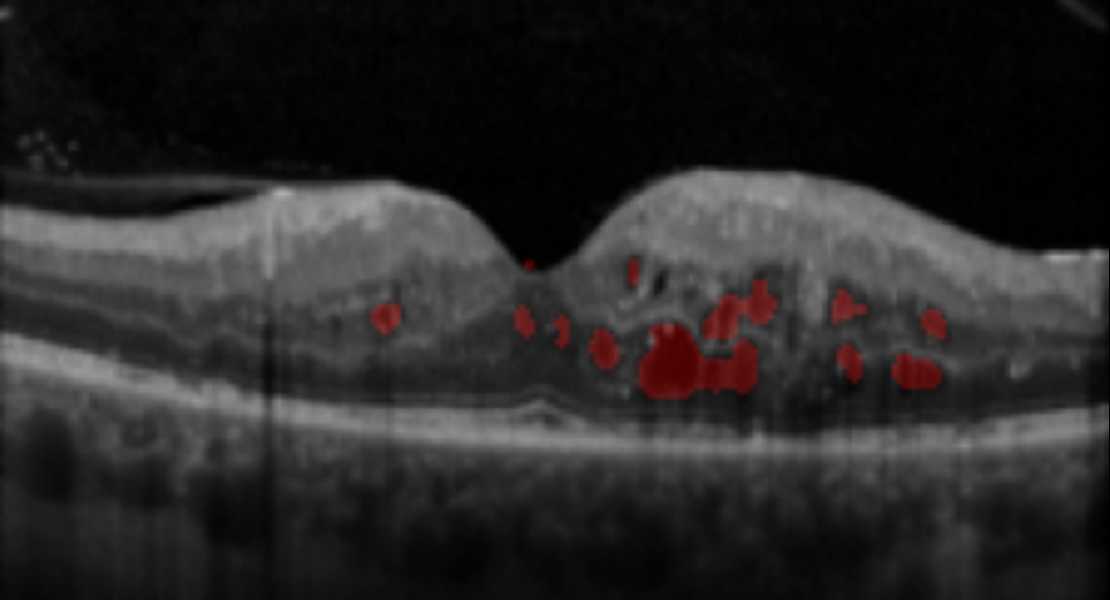} &
\includegraphics[width=1.8cm,height=1.8cm,keepaspectratio]{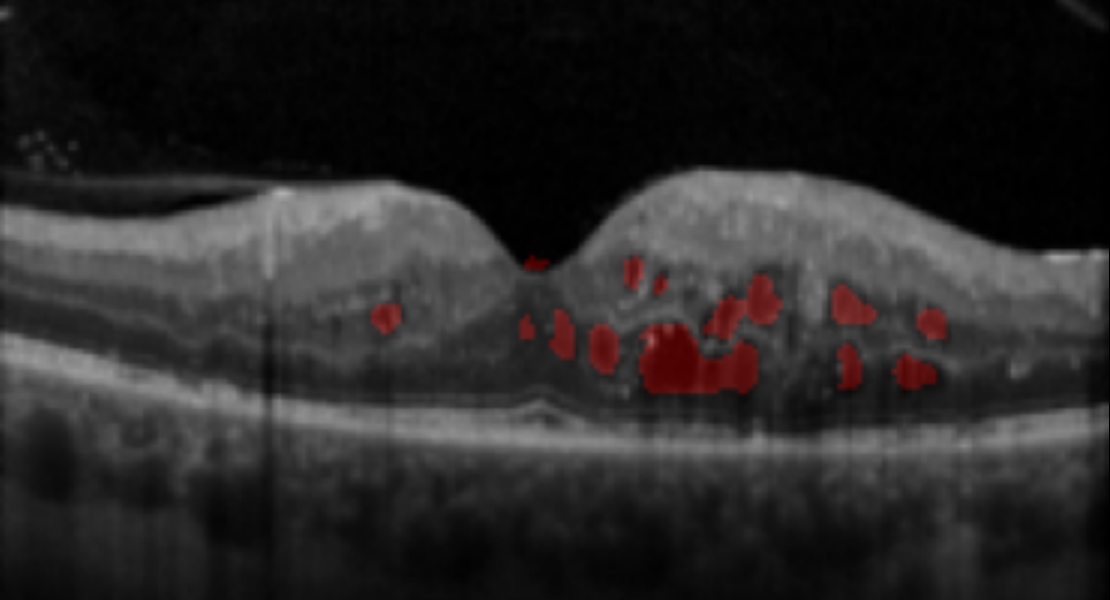} &
\includegraphics[width=1.8cm,height=1.8cm,keepaspectratio]{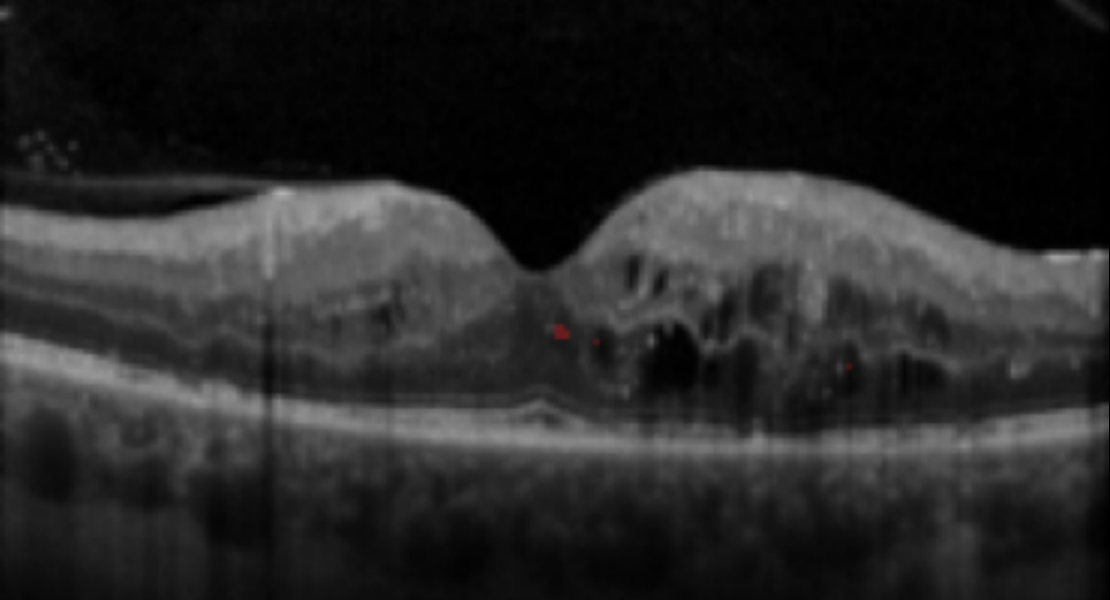} \\
\end{tabular}
}

\caption{
Visual results of clipping strategies. 
The top block shows a COVID-19 sample, while the bottom block shows samples from Duke and UMN. 
Unless otherwise specified, results correspond to UNet++.
OCT images are cropped to remove most of the background layer.
}
\label{fig:qualitative_results_combined}
\end{figure}

\subsection{Impact of clipping strategies on model gradients}
To analyze the effect of clipping on gradient transformation, we measure the alignment between transformed and true batch-averaged gradients using cosine similarity, following \cite{xia2023differentiallya}, and visualize their distributions across datasets and models (Figure~\ref{fig:gradients_all}).
Overall, all methods show predominantly positive cosine similarities concentrated near 1, indicating that gradient direction is broadly preserved under all clipping strategies.
The primary distinction between methods lies in the spread and concentration of these distributions. 
\ac{NSGD} consistently produces the most concentrated distributions near 1, indicating the strongest preservation of gradient direction.
Flat clipping follows closely, while \ac{Auto-S} shows a broader spread, suggesting weaker alignment. 
In contrast, \ac{PSAC} consistently shows the widest distributions, with noticeable mass at lower cosine values, indicating greater directional distortion relative to the true gradient.
These trends are consistent across datasets and models. 


This empirical ordering contrasts with the findings of \cite{xia2023differentiallya}, which report that \ac{PSAC} achieves superior alignment by shifting the cosine similarity distribution closer to 1, while \ac{Auto-S} and \ac{NSGD} exhibit larger deviations. 
In our experiments, however, \ac{PSAC} consistently shows the largest deviation, whereas \ac{NSGD} achieves the strongest alignment. 

A likely explanation is the difference in task structure. 
Prior work focuses on classification tasks, where small gradients are often treated as uninformative noise. 
In contrast, segmentation involves dense prediction, where small gradients, particularly around boundaries and fine structures, can carry meaningful information. 
As a result, re-weighting schemes such as \ac{PSAC}, designed to suppress small gradients, may be less effective in this setting.

Finally, stronger gradient alignment does not necessarily translate to better segmentation performance (Figure~\ref{fig:metrics_duke}--\ref{fig:metrics_lunct}), indicating a decoupling between gradient direction preservation and downstream utility.

\begin{figure}[!t]
\centering

\begin{subfigure}{0.32\linewidth}
  \includegraphics[width=\linewidth]{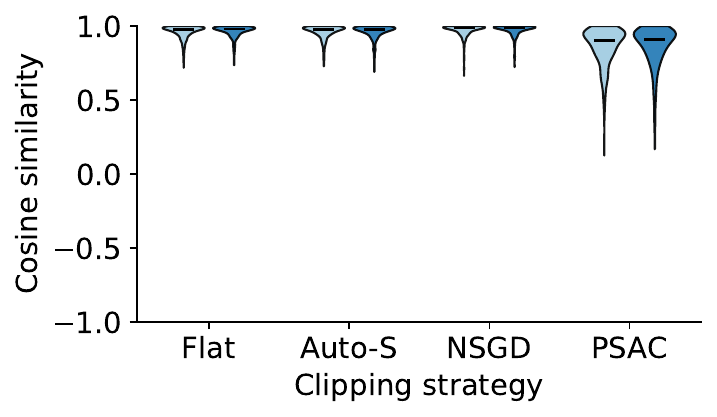}
  \caption{Duke, UNet++}
  \label{duke_gradient}
\end{subfigure}
\hfill
\begin{subfigure}{0.32\linewidth}
  \includegraphics[width=\linewidth]{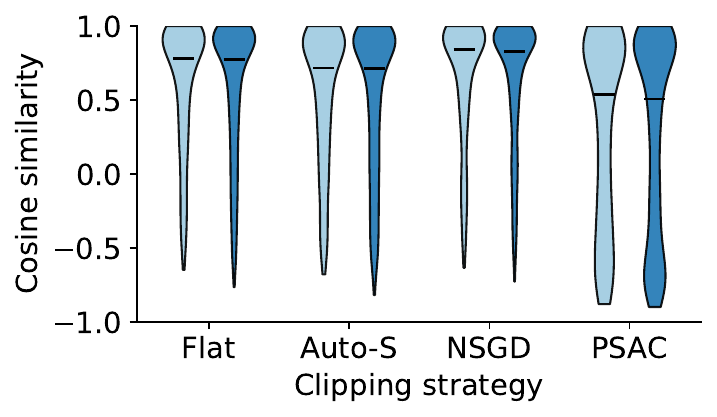}
  \caption{UMN, UNet++}
  \label{umn_gradient}
\end{subfigure}
\hfill
\begin{subfigure}{0.32\linewidth}
  \includegraphics[width=\linewidth]{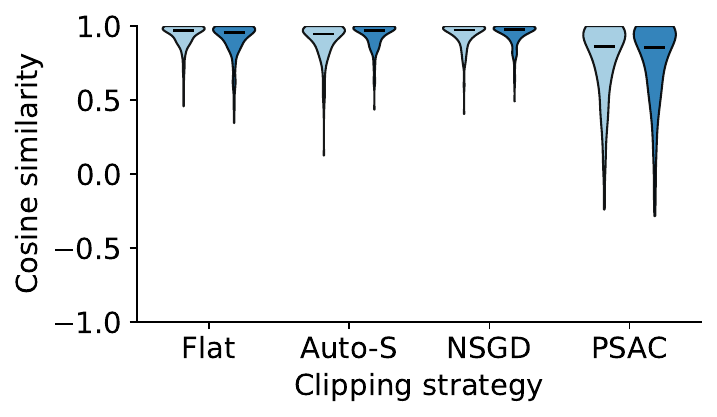}
  \caption{COVID-19, UNet++}
  \label{covid_gradient}
\end{subfigure}

\vspace{0.5em}

\begin{subfigure}{0.32\linewidth}
  \includegraphics[width=\linewidth]{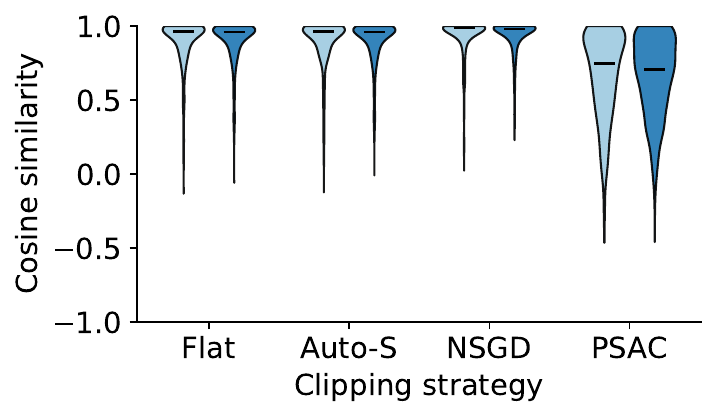}
  \caption{Duke, UNet}
  \label{duke_gradient_2}
\end{subfigure}
\hfill
\begin{subfigure}{0.32\linewidth}
  \includegraphics[width=\linewidth]{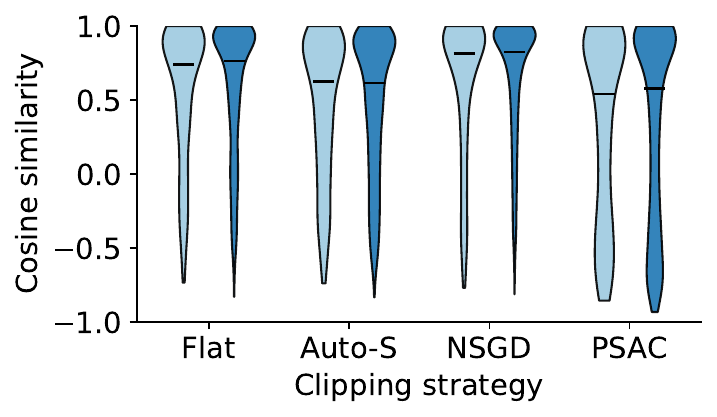}
  \caption{UMN, UNet}
  \label{umn_gradient2}
\end{subfigure}
\hfill
\begin{subfigure}{0.32\linewidth}
  \includegraphics[width=\linewidth]{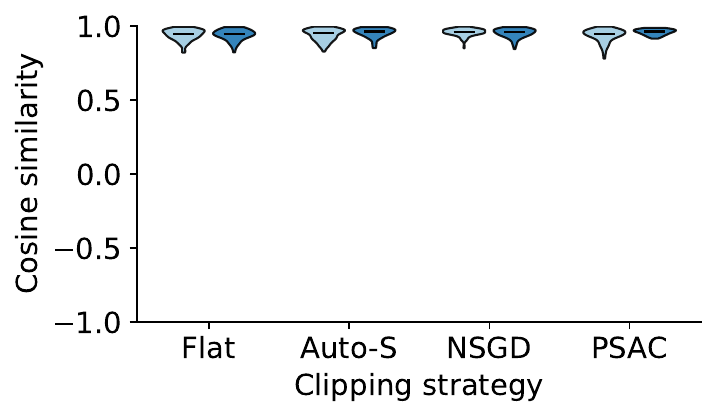}
  \caption{COVID-19, Inf-Net}
  \label{covid_gradient2}
\end{subfigure}

\caption{
Gradient cosine similarity distributions under DP training ($\epsilon=200$).
Top row: UNet++ across datasets. Bottom row: UNet / Inf-Net across datasets.
}
\label{fig:gradients_all}
\end{figure}

\subsection{Computational performance}
To assess the impact of clipping strategies and morphology on training time, we report results for Duke, UMN, and COVID-19 datasets in Figure~\ref{fig:three_models_time}.
Training time is mainly dataset-dependent, with morphology introducing additional overhead.
Among clipping methods, \ac{NSGD} is consistently the most time-efficient (notably on Duke), while \ac{Auto-S} and \ac{PSAC} incur higher computational cost.

\begin{figure}[!t]
  \centering

  \begin{subfigure}[t]{0.32\textwidth}
    \centering
    \includegraphics[width=\linewidth]{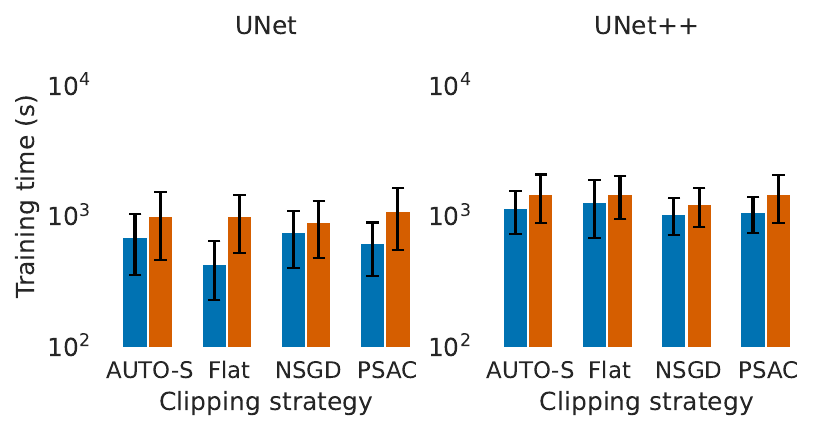}
    \caption{Duke}
    \label{fig:sub1}
  \end{subfigure}
  \hfill
  \begin{subfigure}[t]{0.32\textwidth}
    \centering
    \includegraphics[width=\linewidth]{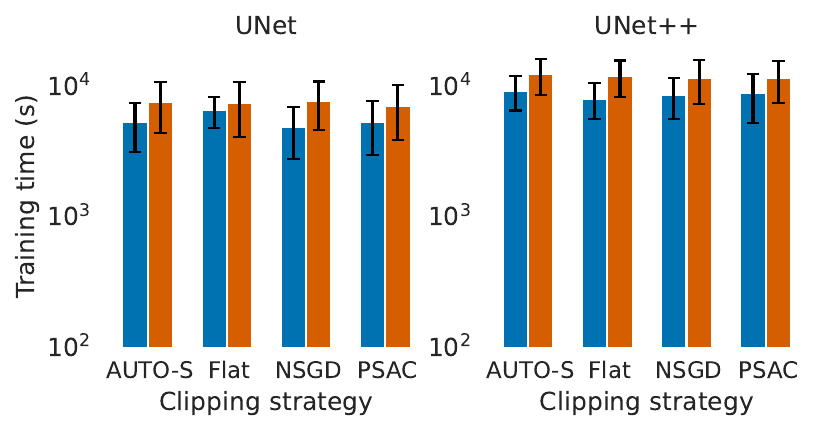}
    \caption{UMN}
    \label{fig:sub2}
  \end{subfigure}
  \hfill
  \begin{subfigure}[t]{0.32\textwidth}
    \centering
    \includegraphics[width=\linewidth]{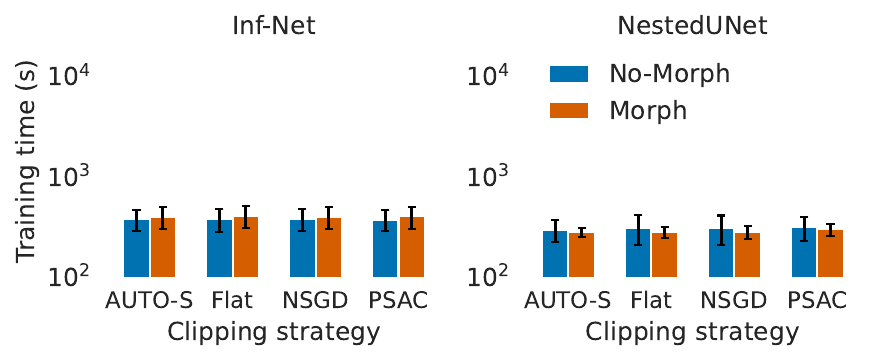}
    \caption{Covid-19 lung CT}
    \label{fig:sub3}
  \end{subfigure}

  \caption{Comparison of training time between different models and clipping strategies in both non-morph and DP-Morph settings for datasets }
  \label{fig:three_models_time}
\end{figure}

\subsection{Impact of clipping strategies on global loss attack}
To gain insight into how different clipping strategies behave under an attack scenario for medical image segmentation models, we analyze their vulnerability to a global-loss attack across datasets and architectures~\cite{chobola2023membership}.
On Duke, \ac{dpsgd} significantly reduces attack success (from $\sim$0.7--0.8 to $<0.5$), with Flat and \ac{Auto-S} slightly outperforming \ac{NSGD} and \ac{PSAC}. 
On UMN, attack success remains above random ($>0.5$), indicating only partial mitigation, with \ac{NSGD} performing best and \ac{PSAC} worst.

For COVID-19, Inf-Net achieves lowest attack success with \ac{Auto-S}/\ac{NSGD} ($\approx 0.50$) and highest with \ac{PSAC} ($\approx 0.62$), while NestedUNet performs best with \ac{Auto-S} ($\approx 0.42$) and worst with \ac{PSAC} or no privacy ($\approx 0.52$).

Overall, clipping strategy has a moderate, model-dependent impact: \ac{Auto-S} generally provides stronger protection, while \ac{PSAC} is weaker.

\begin{figure}[!t]
  \centering

  \begin{subfigure}[t]{0.32\textwidth}
    \centering
    \includegraphics[width=\linewidth]{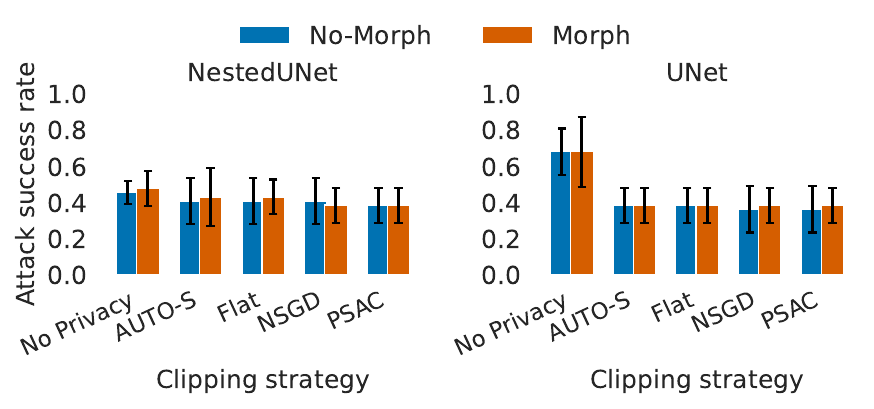}
    \caption{Duke}
    \label{fig:sub1_loss}
  \end{subfigure}
  \hfill
  \begin{subfigure}[t]{0.32\textwidth}
    \centering
    \includegraphics[width=\linewidth]{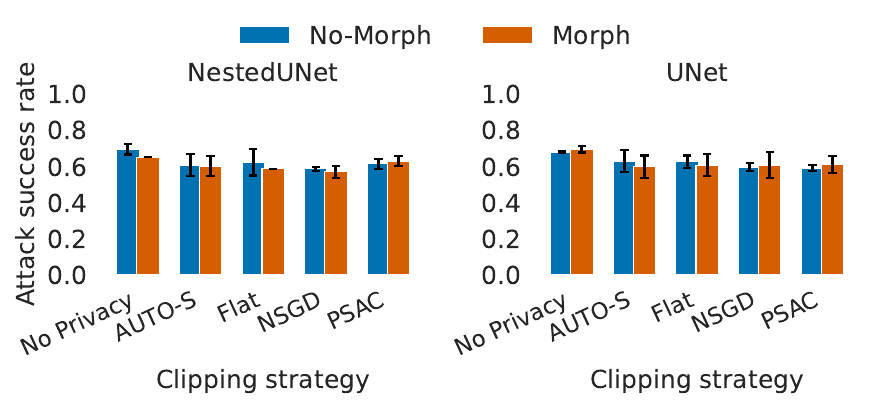}
    \caption{UMN}
    \label{fig:sub2_loss}
  \end{subfigure}
  \hfill
  \begin{subfigure}[t]{0.32\textwidth}
    \centering
    \includegraphics[width=\linewidth]{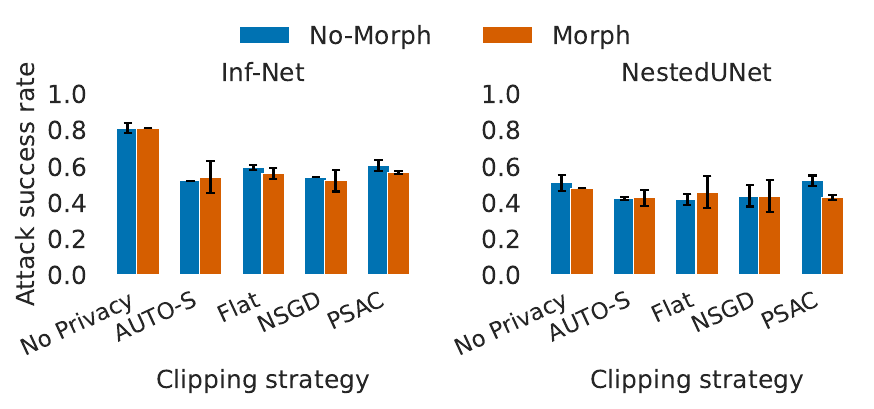}
    \caption{Covid-19 lung CT}
    \label{fig:sub3_loss}
  \end{subfigure}

  \caption{Protection provided by clipping strategies against global loss attack}
  \label{fig:global_loss}
\end{figure}

\subsection{Impact of morphological refinement}

Figure~\ref{fig:oct-inter} shows the effect of morphological operations on segmentation outputs. 
Changes are confined to a small subset of pixels, mainly along layer boundaries, while most regions remain unaffected.
Improvements are concentrated in structurally meaningful regions, indicating that morphology acts as a targeted refinement rather than a global transformation, correcting local inconsistencies without altering well-predicted areas.
This behavior motivates our novel layer-specific morphology, because different layers show distinct structural properties.

\begin{figure}[!t]
\centering
\includegraphics[width=1\linewidth]{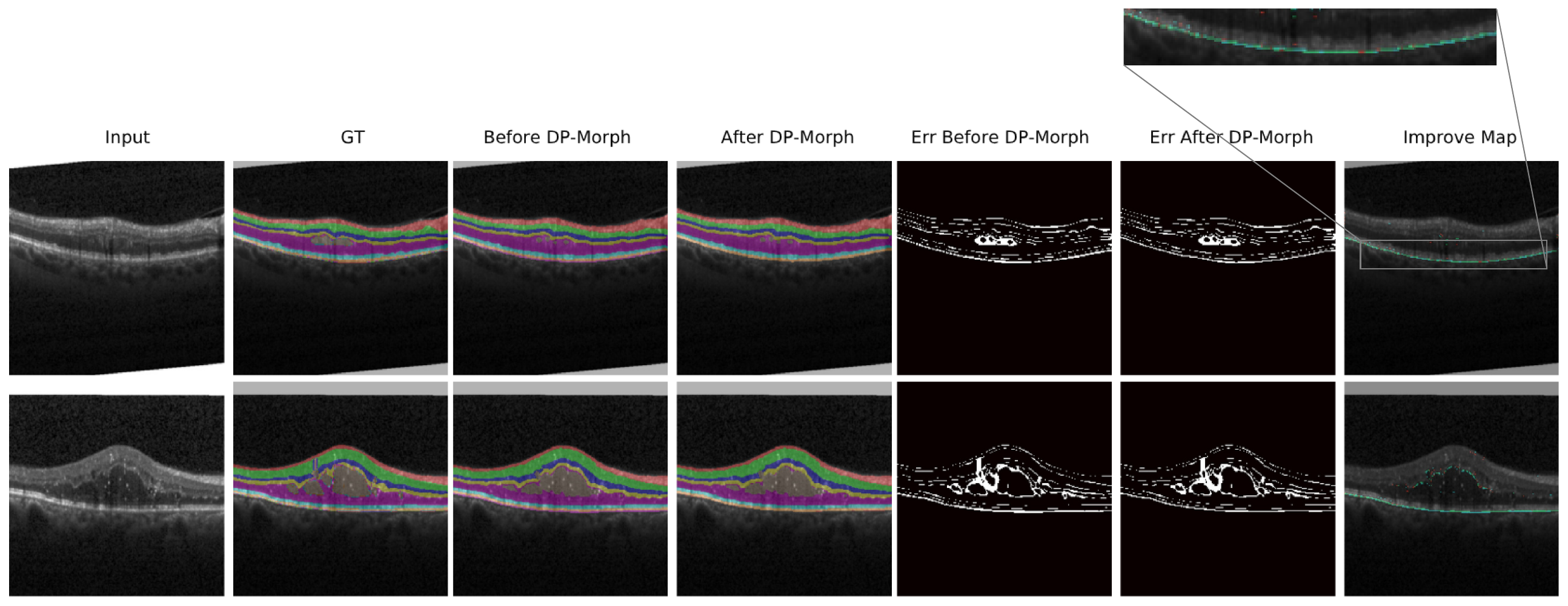}
\caption{Effect of morphological refinement on segmentation predictions. Each row shows the input, ground truth (GT), and model predictions before and after applying morphology. The Improved Map visualizes corrected regions (Cyan) and newly introduced errors (red).}
\label{fig:oct-inter}
\end{figure}

\subsection{Impact of adaptive morphological refinement}


We investigate the impact of adaptive DP-Morph on Duke dataset.
Figure~\ref{fig:adaptive_performance} compares adaptive morphology under policies v1-v3 for UNet++. 
For UNet++, policy v2 provides a balanced behavior, closely matching v1 while avoiding the more aggressive treatment of thick layers in v3. 
In the non-private setting, all models show modest improvements, with gains of up to 1.5\% in Dice. 

The evolution of estimated layer thickness and layer-wise Dice scores (shown for UNet++ under v1 in Figure~\ref{fig:policy_nested}) indicates that adaptive morphology mainly benefits specific layers, rather than uniformly improving all classes.
The effectiveness of adaptive morphology is influenced by the clipping strategy. 
Flat and \ac{Auto-S} preserve a more consistent layer-wise profile, achieving strong performance on layers 1, 2, 5, and 6, while maintaining comparatively better robustness on the more challenging layers under \ac{dpsgd}.
In contrast, \ac{NSGD} and \ac{PSAC} consistently show lower performance, and the introduction of morphology does not substantially mitigate this gap.
The weaker performance of \ac{NSGD} and \ac{PSAC} is not only reflected in lower average Dice, but also in pronounced degradation on more challenging layers, particularly layers 3 and 4. 

Overall, these results show that adaptive morphology is an effective strategy for improving segmentation performance. 
Nevertheless, the layer-wise analysis reveals that certain challenging layers, particularly layers 3 and 4, remain difficult to segment under \ac{dpsgd}.

\begin{figure}[!t]
    \centering
    \includegraphics[width=1\linewidth]{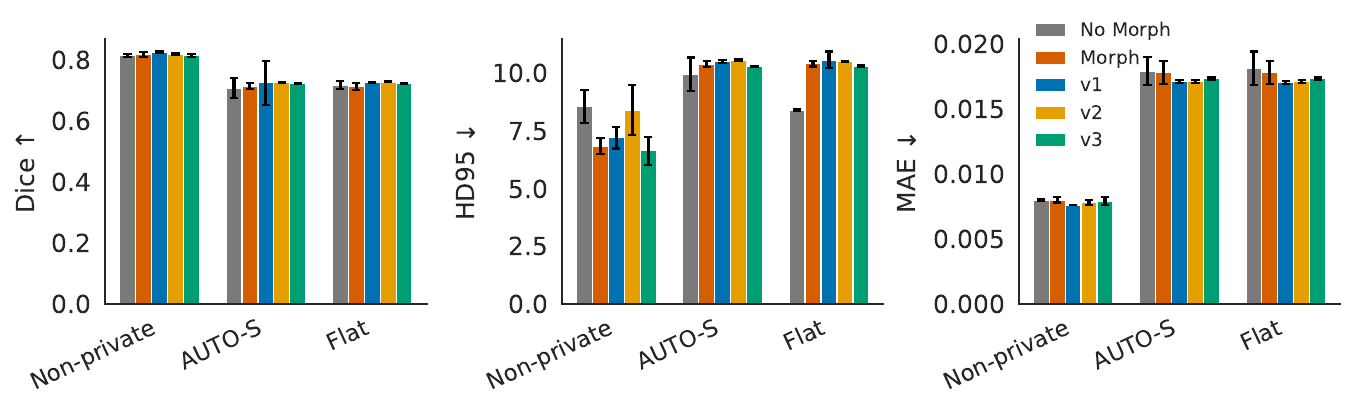}
    \caption{Segmentation performance of adaptive morphology (v1-v3) compared to fixed DP-Morph and no morphology under non-private, Flat, and AUTO-S clipping settings.}
    \label{fig:adaptive_performance}
  \vspace{0.5em}
\end{figure}

\begin{figure}[!t]
    \centering
    \includegraphics[width=1\linewidth]{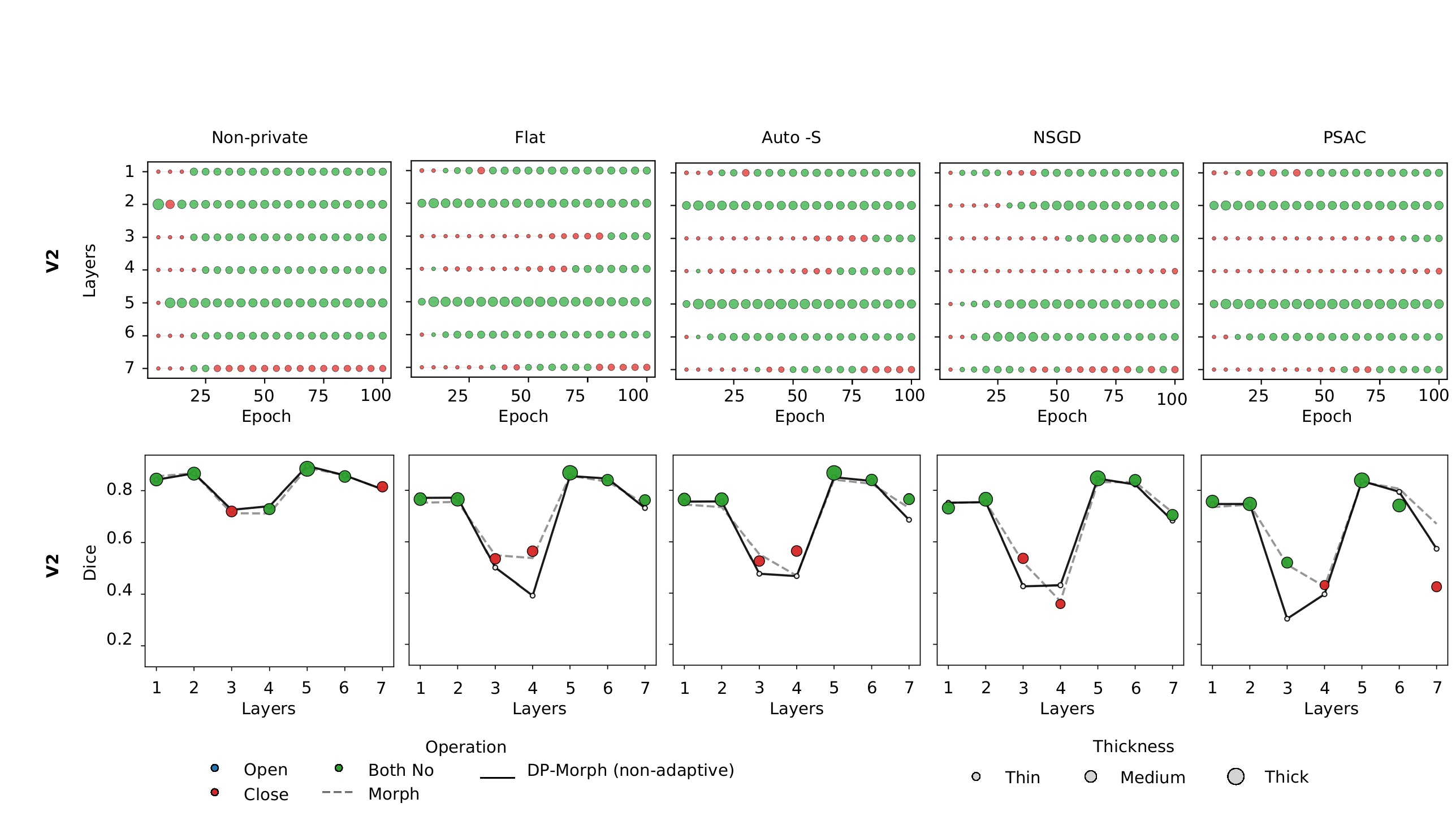}
    \caption{Layer-wise policy evolution during training (first row), and Layer-wise performance and operation assignment under adaptive morphology (second row) with V2 policy for UNet++ on the Duke dataset }
    \label{fig:policy_nested}
  \vspace{0.5em}
\end{figure}

\section{Discussion and Conclusion}

Our results show that the effectiveness of clipping strategies depends on the dataset, model architecture, and training setting. 
\ac{Auto-S} consistently provides stable performance, while \ac{NSGD} often underperforms due to aggressive normalization.
Although \ac{PSAC} aims to better approximate true gradients, its advantages do not consistently transfer to segmentation.
We further observe that morphological refinement interacts non-trivially with clipping strategies, highlighting the importance of jointly considering optimization and structural design. 
Adaptive morphology improves over fixed operations, but challenges remain for difficult layers.
A promising direction is to use topology-aware signals to guide the local refinement of morphological operations, enabling targeted correction of structural errors, potentially leading to improved performance under \ac{dpsgd}.

\bibliographystyle{splncs04}
\bibliography{OCT_Ref}

\appendix
\section*{Appendix}

Additional figures illustrating the evolution of estimated layer thickness and layer-wise Dice scores for UNet++, LFUNet, and UNet across all policies and clipping strategies are available in our Git repository: \url{https://gitlab.com/dmi-pet-public/parsarad2026medicalsegmentationprivay}.

\begin{table}[htbp]
\centering
\caption{Difference in Validation Dice ($\Delta$) relative to $(q_{\text{low}}=20, q_{\text{high}}=85)$ for NestedUNet across policies (v1–v3) on the Duke dataset with $\texttt{retinal\_layer\_wise=True}$. 
We report $\Delta_{25,80} = \text{Dice}_{20,85} - \text{Dice}_{25,80}$ (mean over top-3 runs) and $\Delta_{30,70} = \text{Dice}_{20,85} - \text{Dice}_{30,70}$. 
Positive values indicate improvement when using $(20,85)$.}
\label{tab:quantile_comparison_nestedunet_all_policies}
\begin{tabular}{ll|ccccc}
\toprule
Policy & Setting & Auto-S & Flat & NSGD & PSAC & None \\
\midrule
\multirow{2}{*}{v1} 
 & $\Delta_{25,80}$ & +0.0058 & +0.0039 & +0.0230 & +0.0313 & +0.0231 \\
 & $\Delta_{30,70}$ & -0.0018 & -0.0014 & +0.0216 & +0.0087 & +0.0049 \\
\midrule
\multirow{2}{*}{v2} 
 & $\Delta_{25,80}$ & +0.0010 & +0.0033 & +0.0215 & +0.0310 & +0.0054 \\
 & $\Delta_{30,70}$ & +0.0028 & +0.0036 & -0.0014 & +0.0002 & 0.0000 \\
\midrule
\multirow{2}{*}{v3} 
 & $\Delta_{25,80}$ & +0.0068 & +0.0081 & +0.0088 & +0.0086 & +0.0010 \\
 & $\Delta_{30,70}$ & +0.0140 & +0.0150 & 0.0000 & +0.0134 & 0.0000 \\
\bottomrule
\end{tabular}
\end{table}

\begin{table}[ht!]
\centering
\caption{Best results for UNet++, UNet and LFUNet on the Duke dataset.}
\label{tab:combined_models_best_duke}
\resizebox{\textwidth}{0.59\textheight}{%
\begin{tabular}{l l l|c c c}
\toprule
Model & Setting & Method & Dice $\uparrow$ & \ac{HD95} $\downarrow$ & MAE $\downarrow$ \\
\midrule
\multicolumn{6}{c}{\textbf{UNet++}} \\
\midrule
\multirow{5}{*}{UNet++} & \multirow{5}{*}{Non-private} & No Morph & 0.8145 $\pm$ 0.0047 & 8.5458 $\pm$ 0.7239 & 0.0080 $\pm$ 0.0001 \\
& & Morph (both) & 0.8177 $\pm$ 0.0083 & 6.8436 $\pm$ 0.3423 & 0.0080 $\pm$ 0.0002 \\
& & Adaptive Morph (v1) & \textbf{0.8266 $\pm$ 0.0019}& 7.2070 $\pm$ 0.4695& \textbf{0.0076$\pm$ 0.0000} \\
& & Adaptive Morph (v2) & 0.8203 $\pm$ 0.0038 & 8.4047$\pm$ 1.0660 & 0.0078 $\pm$ 0.0002\\
& & Adaptive Morph (v3) & 0.8164 $\pm$ 0.0048  & \textbf{6.6387 $\pm$ 0.6022}&0.0079 $\pm$ 0.0003 \\
\cmidrule(lr){2-6}
\multirow{4}{*}{UNet++} & \multirow{4}{*}{DP (no morph)} & AUTO-S & 0.7088 $\pm$ 0.0317 & 9.9473 $\pm$ 0.7297 & 0.0179 $\pm$ 0.0011 \\
& & Flat & 0.7187 $\pm$ 0.0140 & \textbf{8.4079 $\pm$ 0.0552} & 0.0181 $\pm$ 0.0013 \\
& & NSGD & 0.6840 $\pm$ 0.0143 & 9.8162 $\pm$ 1.1241 & 0.0209 $\pm$ 0.0002 \\
& & PSAC & 0.6735 $\pm$ 0.0269 & 9.3752 $\pm$ 2.0165 & 0.0223 $\pm$ 0.0000 \\
\cmidrule(lr){2-6}
\multirow{4}{*}{UNet++} & \multirow{4}{*}{DP + Morph} & AUTO-S & 0.7153 $\pm$ 0.0093 & 10.3826 $\pm$ 0.1275 & 0.0178 $\pm$ 0.0009 \\
& & Flat & 0.7129 $\pm$ 0.0109 & 10.4055 $\pm$ 0.1157 & 0.0178 $\pm$ 0.0009 \\
& & NSGD & 0.6736 $\pm$ 0.0407 & 10.8702 $\pm$ 0.5375 & 0.0211 $\pm$ 0.0008 \\
& & PSAC & 0.6483 $\pm$ 0.0213 & 10.1481 $\pm$ 0.3806 & 0.0227 $\pm$ 0.0007 \\
\cmidrule(lr){2-6}
\multirow{4}{*}{UNet++} & \multirow{4}{*}{DP + Adaptive Morph (v1)} & AUTO-S & 0.7257$\pm$ 0.0723  & 10.5094$\pm$ 0.0723 & \textbf{0.0171$\pm$ 0.0001} \\
& & Flat & 0.7271$\pm$ 0.0018 & 10.5740$\pm$ 0.3641 & 0.0170$\pm$ 0.0001 \\
& & NSGD & 0.6740$\pm$ 0.0106 & 10.7742$\pm$ 0.2686 & 0.0200 $\pm$ 0.0008\\
& & PSAC & 0.6517 $\pm$ 0.0075& 10.9951$\pm$ 0.4056 & 0.0224$\pm$ 0.0014 \\
\cmidrule(lr){2-6}
\multirow{4}{*}{UNet++} & \multirow{4}{*}{DP + Adaptive Morph (v2)} & AUTO-S & 0.7276$\pm$ 0.0015  & 10.5607 $\pm$ 0.0354& \textbf{0.0171 $\pm$ 0.0001}\\
& & Flat & \textbf{0.7286$\pm$ 0.0015} & 10.5057 $\pm$ 0.0335 & \textbf{0.0171  $\pm$ 0.0001} \\
& & NSGD & 0.6830 $\pm$ 0.0032 & 10.8621 $\pm$ 0.1024& 0.0191$\pm$ 0.0005  \\
& & PSAC & 0.6427$\pm$ 0.0030 & 12.1029$\pm$ 0.7417 & 0.0200 $\pm$ 0.0016\\
\cmidrule(lr){2-6}
\multirow{4}{*}{UNet++} & \multirow{4}{*}{DP + Adaptive Morph (v3)} & AUTO-S & 0.7229 $\pm$ 0.0012 & 10.2955 $\pm$ 0.0114 & 0.0173 $\pm$ 0.0001 \\
& & Flat & 0.7244 $\pm$ 0.0020 & 10.3048 $\pm$ 0.0255 & 0.0173 $\pm$ 0.0001 \\
& & NSGD & 0.7266 $\pm$ 0.0018 & 10.3161 $\pm$ 0.0444 & \textbf{0.0171 $\pm$ 0.0001} \\
& & PSAC & 0.6388 $\pm$ 0.0023 & 13.1750$\pm$ 0.3744 & 0.0198 $\pm$ 0.0001 \\
\midrule

\multicolumn{6}{c}{\textbf{UNet}} \\
\midrule
\multirow{5}{*}{UNet} & \multirow{5}{*}{Non-private} & No Morph & 0.8158 $\pm$ 0.0057 & 8.1407 $\pm$ 0.5123 & 0.0080 $\pm$ 0.0002 \\
& & Morph (both) & 0.8218 $\pm$ 0.0026 & 6.3642 $\pm$ 0.3780 & 0.0079 $\pm$ 0.0001 \\
& & Adaptive Morph (v1) & \textbf{0.8266 $\pm$ 0.0018} & 8.5204$\pm$ 1.1475 & 0.0077 $\pm$ 0.0002 \\
& & Adaptive Morph (v2)& 0.8240 $\pm$ 0.0016 & 6.7208 $\pm$ 0.4319 & \textbf{0.0076 $\pm$ 0.0001} \\
& & Adaptive Morph (v3) & 0.8225 $\pm$ 0.0048 & \textbf{6.2871} $\pm$ 0.9231 & 0.0078 $\pm$0.0002 \\
\cmidrule(lr){2-6}
\multirow{4}{*}{UNet} & \multirow{4}{*}{DP (no morph)} & AUTO-S & 0.6579 $\pm$ 0.0821 & \textbf{8.0900 $\pm$ 0.1607} & 0.0221 $\pm$ 0.0017 \\
& & Flat & 0.7181 $\pm$ 0.0253 & 8.5468 $\pm$ 0.1564 & 0.0187 $\pm$ 0.0011 \\
& & NSGD & 0.4727 $\pm$ 0.1079 & 13.9979 $\pm$ 0.8943 & 0.0267 $\pm$ 0.0021 \\
& & PSAC & 0.6637 $\pm$ 0.0491 & 9.7710 $\pm$ 1.1850 & 0.0214 $\pm$ 0.0020 \\
\cmidrule(lr){2-6}
\multirow{4}{*}{UNet} & \multirow{4}{*}{DP + Morph} & AUTO-S & \textbf{0.7240 $\pm$ 0.0255} & 9.7290 $\pm$ 0.3777 & 0.0181 $\pm$ 0.0011 \\
& & Flat & 0.6945 $\pm$ 0.0085 & 9.7305 $\pm$ 0.2718 & \textbf{0.0178 $\pm$ 0.0010} \\
& & NSGD & 0.6140 $\pm$ 0.0599 & 11.5495 $\pm$ 2.6518 & 0.0220 $\pm$ 0.0026 \\
& & PSAC & 0.6687 $\pm$ 0.0253 & 9.3020 $\pm$ 0.6968 & 0.0276 $\pm$ 0.0073 \\
\cmidrule(lr){2-6}
\multirow{4}{*}{UNet} & \multirow{4}{*}{DP + Adaptive Morph (v1)} & AUTO-S & 0.6877$\pm$ 0.0499 & 8.6223$\pm$ 0.1548 & 0.0178 $\pm$ 0.0021\\
& & Flat & 0.6910 $\pm$ 0.0148 & 8.6912 $\pm$ 0.6529 & 0.0177 $\pm$ 0.0003 \\
& & NSGD & 0.5392 $\pm$ 0.0179 & 11.4330 $\pm$ 2.7424& 0.0236 $\pm$ 0.0008\\
& & PSAC & 0.6185 $\pm$ 0.0113  & 10.1166 $\pm$ 0.9327& 0.0221 $\pm$ 0.0004\\
\midrule
\cmidrule(lr){2-6}
\multirow{4}{*}{UNet} & \multirow{4}{*}{DP + Adaptive Morph (v2)} & AUTO-S & 0.6934 $\pm$ 0.0045 & 8.3737$\pm$ 0.3021 & 0.0182 $\pm$ 0.0003  \\
& & Flat & 0.6929 $\pm$ 0.0079 & 8.5344 $\pm$ 0.7502 & 0.0181 $\pm$ 0.0003\\
& & NSGD & 0.5398 $\pm$ 0.0079 & 17.0538 $\pm$ 1.9071 & 0.0234 $\pm$ 0.0000  \\
& & PSAC & 0.6211 $\pm$ 0.0096  & 8.5174 $\pm$ 1.2708 & 0.0222 $\pm$ 0.0004 \\
\cmidrule(lr){2-6}
\multirow{4}{*}{UNet} & \multirow{4}{*}{DP + Adaptive Morph (v3)} & AUTO-S & 0.6832 $\pm$ 0.0050 & 9.1165 $\pm$ 0.2417 & 0.0181 $\pm$ 0.0002 \\
& & Flat & 0.6851 $\pm$ 0.0049 & 9.0948 $\pm$ 0.2694 & 0.0181 $\pm$ 0.0002 \\
& & NSGD & 0.5386 $\pm$ 0.0142 & 13.5182 $\pm$ 1.3098 & 0.0230 $\pm$ 0.0004 \\
& & PSAC & 0.5767 $\pm$ 0.0095 & 15.2210 $\pm$ 3.6294 & 0.0224 $\pm$ 0.0006 \\
\midrule
\multicolumn{6}{c}{\textbf{LFUNet}} \\
\midrule
\multirow{5}{*}{LFUNet} & \multirow{5}{*}{Non-private} & No Morph & 0.8119 $\pm$ 0.0050 & \textbf{5.7770 $\pm$ 0.5183} & 0.0080 $\pm$ 0.0003 \\
& & Morph (both) & 0.8050 $\pm$ 0.0092 & 6.5911 $\pm$ 0.4710 & 0.0084 $\pm$ 0.0004 \\
& & Adaptive Morph (v1) & 0.8144 $\pm$ 0.0006 & 6.4272 $\pm$ 0.1923 & 0.0079 $\pm$ 0.0001  \\
& & Adaptive Morph (v2) & 0.8130 $\pm$ 0.0005 & 6.0715 $\pm$ 0.4405  & 0.0080 $\pm$ 0.0001 \\
& & Adaptive Morph (v3) & \textbf{0.8172 $\pm$ 0.0113} & 6.4908 $\pm$ 0.3108& \textbf{0.0078 $\pm$ 0.0005} \\
\cmidrule(lr){2-6}
\multirow{4}{*}{LFUNet} & \multirow{4}{*}{DP (no morph)} & AUTO-S & 0.6886 $\pm$ 0.0011 & 10.5084 $\pm$ 0.0843 & 0.0146 $\pm$ 0.0002 \\
& & Flat & 0.6857 $\pm$ 0.0041 & \textbf{10.2825 $\pm$ 0.1663} & 0.0145 $\pm$ 0.0003 \\
& & NSGD & 0.6700 $\pm$ 0.0061 & 10.3767 $\pm$ 0.0071 & 0.0156 $\pm$ 0.0000 \\
& & PSAC & 0.6773 $\pm$ 0.0035 & 10.5747 $\pm$ 0.1268 & 0.0148 $\pm$ 0.0005 \\
\cmidrule(lr){2-6}
\multirow{4}{*}{LFUNet} & \multirow{4}{*}{DP + Morph} & AUTO-S & 0.6825 $\pm$ 0.0153 & 10.5554 $\pm$ 0.1523 & 0.0146 $\pm$ 0.0007 \\
& & Flat & 0.6860 $\pm$ 0.0115 & 10.6573 $\pm$ 0.1890 & 0.0146 $\pm$ 0.0004 \\
& & NSGD & 0.6626 $\pm$ 0.0168 & 10.8570 $\pm$ 0.4087 & 0.0162 $\pm$ 0.0004 \\
& & PSAC & 0.6674 $\pm$ 0.0030 & 10.5192 $\pm$ 0.5173 & 0.0154 $\pm$ 0.0003 \\
\cmidrule(lr){2-6}
\multirow{4}{*}{LFUNet} & \multirow{4}{*}{DP + Adaptive Morph (v1)} & AUTO-S & 0.6961$\pm$ 0.0006 & 10.5679$\pm$ 0.0766 & 0.0141$\pm$ 0.0003 \\
& & Flat & 0.6964$\pm$ 0.0007 & 10.6008 $\pm$ 0.0619 & 0.0141$\pm$ 0.0003 \\
& & NSGD & 0.6784  $\pm$ 0.0006& 10.7826 $\pm$ 0.0999& 0.0160 $\pm$ 0.0002\\
& & PSAC & 0.6726 $\pm$ 0.0012 & 10.6323 $\pm$ 0.3226  & 0.0157 $\pm$ 0.0001\\
\cmidrule(lr){2-6}
\multirow{4}{*}{LFUNet} & \multirow{4}{*}{DP + Adaptive Morph (v2)} & AUTO-S & 0.6961 $\pm$ 0.0006& 10.5679 $\pm$ 0.1064 & \textbf{0.0141 $\pm$ 0.0001}\\
& & Flat & \textbf{0.6964 $\pm$ 0.0008}& 10.6008 $\pm$ 0.0884& \textbf{0.0141$\pm$ 0.0001} \\
& & NSGD & 0.6784 $\pm$ 0.0003& 10.7826 $\pm$ 0.1259 & 0.0160$\pm$ 0.0002 \\
& & PSAC & 0.6726 $\pm$ 0.0006 & 10.6323 $\pm$ 0.0794 & 0.0157 $\pm$ 0.0001\\
\cmidrule(lr){2-6}
\multirow{4}{*}{LFUNet} & \multirow{4}{*}{DP + Adaptive Morph (v3)} & AUTO-S & 0.6972 $\pm$ 0.0010 & 10.7540 $\pm$ 0.1117 & 0.0143 $\pm$ 0.0001 \\
& & Flat & 0.6974 $\pm$ 0.0008 & 10.7434 $\pm$ 0.1003 & 0.0143  $\pm$ 0.0001 \\
& & NSGD & 0.6798 $\pm$ 0.0009 & 10.7543 $\pm$ 0.1769 & 0.0160 $\pm$ 0.0002 \\
& & PSAC & 0.6768 $\pm$ 0.0018 & 10.6243 $\pm$ 0.0753 & 0.0156 $\pm$ 0.0001 \\
\bottomrule
\end{tabular}
}
\end{table}

\end{document}